%% file: main_neurips.tex
\documentclass{article}

\PassOptionsToPackage{authoryear}{natbib}
\usepackage[main, final]{neurips_2026}

\usepackage[utf8]{inputenc} %
\usepackage[T1]{fontenc}    %
\usepackage{hyperref}       %
\usepackage{url}            %
\usepackage{booktabs}       %
\usepackage{amsfonts}       %
\usepackage{nicefrac}       %
\usepackage{microtype}      %
\usepackage{lipsum}
\usepackage{graphicx}
\graphicspath{{media/}}     %
\usepackage{enumitem}
\usepackage{amsmath}
\usepackage{subcaption}
\usepackage[font=small,labelfont=bf]{caption}
\usepackage[dvipsnames,table,xcdraw]{xcolor}
\usepackage{longtable}
\usepackage{xspace}
\usepackage{multirow}
\usepackage{wrapfig}
\usepackage{colortbl} 
\usepackage{caption}
\usepackage{adjustbox}

\usepackage{tabularx}
\usepackage{amssymb} %
\usepackage{listings}
\definecolor{perc_y}{HTML}{FFF2CC} %
\definecolor{ind_b}{HTML}{D9E5FF}  %
\definecolor{exp_r}{HTML}{FFD9D9}  %

\newcommand{\cf}{\textit{cf.}~}

\definecolor{darkgreen}{HTML}{006401}   %
\definecolor{darkred}{HTML}{8B0000}     %
\newcommand{\imp}[1]{{\tiny\textcolor{darkgreen}{($\Delta$#1\%)}}}
\newcommand{\impneg}[1]{{\tiny\textcolor{darkred}{($\Delta$#1\%)}}}

\newcommand{\environment}{\mbox{\textsc{ZendoWorld}}\xspace}

\title{Playing ZendoWorld: Challenging AI Agents on Active Visual Concept Induction}
\author{Sophia Koehler$^{1, 2}$
\and
\textbf{Antonia Wüst}$^{1}$
\and
\textbf{Inga Ibs}$^{3, 4}$
\and
\textbf{Wasu Top Piriyakulkij}$^{5}$
\and
\textbf{Wolfgang Stammer}$^{6}$
\and
\textbf{Constantin Rothkopf}$^{2, 3, 4}$
\and 
\textbf{Kevin Ellis}$^{5}$
\and 
\textbf{Kristian Kersting}$^{1, 2, 4, 7}$
\and
$^{1}$AIML Lab, TU Darmstadt;\\ 
$^{2}$Hessian Center for AI (hessian.AI); 
$^{3}$ Psychology of Information Processing, TU Darmstadt; \\
$^{4}$Centre for Cognitive Science, TU Darmstadt; %
$^{5}$Cornell University;\\
$^{6}$Max Planck Institute for Informatics, Saarland Informatics Campus; %
$^{7}$German Center for AI (DFKI)\\
}
\begin{document}
\maketitle

\begin{abstract}
A central challenge in building intelligent systems is enabling agents to jointly perceive complex inputs, form hypotheses about hidden patterns, and design informative experiments to test them.
To study this problem, we introduce \environment, a controlled interactive environment in which agents must infer latent logical rules governing visual game observations by actively proposing new scenes, gathering feedback from the environment, and iteratively refining their hypotheses.
We evaluate several agents spanning pure VLM reasoning, Bayesian particle filtering, dynamic concept discovery, and neuro‑symbolic methods. Our main findings are: (1) high accuracy in predicting labels for observed examples does not imply recovery of the underlying rule; (2) perception and induction are distinct bottlenecks for different agent classes; and (3) VLM-based agents propose less informative experiments than random scene generation under our information-gain metric.
To compare these results, we collect human data on the game, which reveals a gap in favor of humans in inductive reasoning and experimentation, particularly for more complex rules.
Overall, \environment takes an important step toward evaluating intelligent agents and identifies concrete avenues for improvement, particularly in domains like scientific discovery.\\
\small{\textbf{Code}: \url{https://github.com/ml-research/ZendoWorld}}\\
\small{\textbf{Data}: \url{https://huggingface.co/datasets/sophia1ch/zendo-synthetic-data}}
\end{abstract}

\section{Introduction}
\begin{wrapfigure}{r}{0.4\textwidth}
  \centering
  \vspace{-5pt} %
  \includegraphics[width=0.9\linewidth]{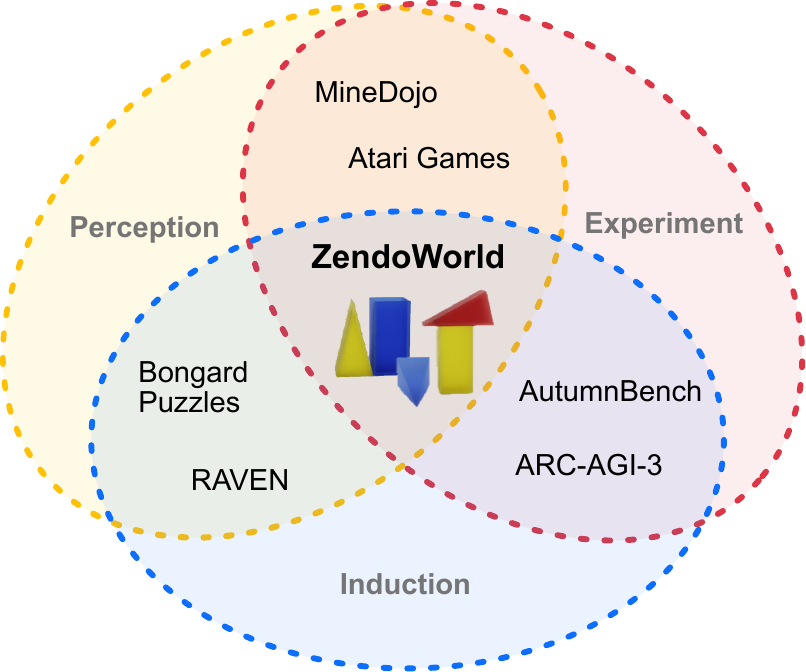}
  \caption{\environment combines perception, induction, and experimentation in a controlled visual environment.}
  \label{fig:motivation}
  \vspace{-15pt} %
\end{wrapfigure}

The ability to form, test, and revise hypotheses in light of new evidence is a hallmark of human intelligence, reflecting the interplay between perception and inductive reasoning. Cognitive science has long emphasized this process as central to learning, showing how humans actively explore hypothesis spaces through informative experiments~\citep{Feist1998ThePO,COOK2011341,oaksford1994rational}. From a causal perspective, this process can be understood as learning not only from observations but from interventions that probe underlying mechanisms~\citep{pearl2009causality, pearl2018book}. With the growing interest in AI agents and AI scientists~\citep{king2009robot, Lu2024TheASA, xie2025far}, structured benchmarks and controlled analyses are needed to assess these capabilities systematically.

Recent progress in AI has focused on inductive reasoning from few-shot examples~\citep{wang2023hypothesis, li2025combining, helff2026slr, wust2025synthesizing}, typically evaluated on grid-based puzzles such as ARC-AGI~\citep{chollet2019measure} or visual reasoning tasks over synthetic and real images~\citep{bongard_pattern_1970, zhang2019raven, jiang2022bongard, wubongard, steinmann2025object} (\cf \autoref{fig:motivation} blue $\cap$ yellow regions). While these benchmarks highlight advances in logical reasoning, particularly in the visual domain, they largely portray induction as a passive, one-off task, thereby overlooking the iterative nature of the discovery process. 

Recent studies have begun to bridge this gap by investigating active experimentation. For instance, \cite{piriyakulkij2024doing} evaluate how LLMs can propose experiments to discover rules expressed in natural language. Yet, in such environments, the data the systems operate on are clean symbolic abstractions, leaving out the complex nature of visual input. Similarly, recent pushes in ARC-AGI-3 \citep{arcagi3_2026} and AutumnBench \citep{warrier2025benchmarking} toward experimentation still rely on input spaces of symbolic nature (\cf \autoref{fig:motivation} blue $\cap$ red).

The more realistic setting lies at the intersection of active experimentation and visual grounding. To study this, we introduce \environment, an interactive benchmark inspired by the inductive logic game \emph{Zendo}\footnote{\url{https://www.looneylabs.com/games/zendo}}. In this multi-turn setting, agents observe labeled scenes, infer a hidden rule, and generate new scenes to test their hypotheses. By combining raw visual input with a formal domain-specific language (DSL) and precise feedback, \environment enables controlled evaluation of perception, induction, and experimentation within a single loop (\cf \autoref{fig:motivation}).

We evaluate agents with different degrees of VLM integration and symbolic structure. The tested methods prove highly specialized: while some excel at labeling or structured search, none reliably solve the full interactive loop without prior knowledge of the rule space. Moreover, labeling accuracy is often dissociated from rule recovery, agents may classify examples correctly while failing to generate informative experiments.

In summary, we (i) introduce \environment, a controlled benchmark for grounded active visual rule discovery spanning 28 games; (ii) develop an evaluation protocol for hypothesis formulation, testing, and revision; (iii) evaluate diverse agent architectures and show that labeling accuracy and rule recovery are strongly dissociated; and (iv) conduct a human study revealing more stable hypothesis trajectories and stronger recovery of complex rules.

\section{Related Work}
\begin{table}[t]
\centering
\caption{Taxonomy of benchmarks at the intersection of perception, reasoning and experimentation.}
\label{tab:benchmark_comparison}
\footnotesize
\begin{tabularx}{\textwidth}{l X c c c}
\toprule
\textbf{Benchmark} & \textbf{Input} & \cellcolor{perc_y}\textbf{Perception} & \cellcolor{ind_b}\textbf{Induction} & \cellcolor{exp_r}\textbf{Experimentation} \\ 
\midrule
ARC-AGI-1/2 & Symbolic (Grid) & \texttimes & \cellcolor{ind_b}\checkmark & \texttimes \\
Bongard, RAVEN & Graphic & \cellcolor{perc_y}\checkmark & \cellcolor{ind_b}\checkmark & \texttimes \\
Bongard-style & Visual (3D) & \cellcolor{perc_y}\checkmark & \cellcolor{ind_b}\checkmark & \texttimes \\
\midrule
Atari, NetHack, MineDojo & Visual & \cellcolor{perc_y}\checkmark & \texttimes\textsuperscript{*} & \cellcolor{exp_r}\checkmark \\
CLEVR-AVR & Visual (3D) & \cellcolor{perc_y}\checkmark & \texttimes\textsuperscript{**} & \cellcolor{exp_r}\checkmark \\
\midrule
Zendo, ActiveACRE & Symbolic (Text) & \texttimes & \cellcolor{ind_b}\checkmark & \cellcolor{exp_r}\checkmark \\
ARC-AGI-3, AutumnBench & Symbolic (Grid) & \texttimes & \cellcolor{ind_b}\checkmark & \cellcolor{exp_r}\checkmark \\
Science-Gym & Symbolic \& Visual (2D) & \cellcolor{perc_y}\checkmark & \cellcolor{ind_b}\checkmark & \cellcolor{exp_r}\checkmark \\
\midrule
\textbf{\environment (Ours)} & \textbf{Visual (3D)} & \cellcolor{perc_y}\checkmark & \cellcolor{ind_b}\checkmark & \cellcolor{exp_r}\checkmark \\ 
\bottomrule
\end{tabularx}
\vspace{0.5em}
\scriptsize{\textsuperscript{*}RL agents typically perform implicit policy optimization rather than explicit hypothesis formulation/rule recovery. \\ \textsuperscript{**}Reasoning over images rather than explicit hypothesis formulation/rule recovery.}
\vspace{-15px}
\end{table}
\textbf{Visual Induction.} A long line of work studies inductive reasoning from fixed visual examples. Classical benchmarks such as Bongard Problems~\citep{bongard_pattern_1970, nie2020bongard}, Raven’s Progressive Matrices~\citep{zhang2019raven}, and ARC-AGI~\citep{chollet2019measure} emphasize abstraction over synthetic or symbolic inputs, while recent Bongard variants extend to natural images and more open-ended concepts~\citep{jiang2022bongard, wubongard, Pawlonka25bongardrwr, malkinski2025reasoning}. As summarized in \autoref{tab:benchmark_comparison}, these settings are largely passive, without allowing agents to query new examples.

Methods in this space fall into three broad classes. First, end-to-end VLMs perform few-shot reasoning directly over images~\citep{johnson2023image, wuest2025bongard}. Second, neuro-symbolic and inductive logic programming approaches recover explicit rules from structured representations: Pix2Code~\citep{wust2024pix2code} extracts objects and synthesizes concepts via DreamCoder-style search~\citep{ellis2023dreamcoder}, while $\alpha$ILP~\citep{shindo2023alpha} models scenes as differentiable logic programs. COMBO~\citep{cropper2023learning} and JOINER~\citep{hocquette2024learning} induce increasingly complex logical rules from fixed symbolic examples, including Zendo structures, but assume predefined symbolic representations rather than reasoning directly over visual inputs. Third, Vision Language Programs (VLP)~\citep{wust2025synthesizing} lift visual inputs into a DSL and synthesize rules over the induced representations.

\begin{figure}[t]
\centering
\includegraphics[width=\linewidth]{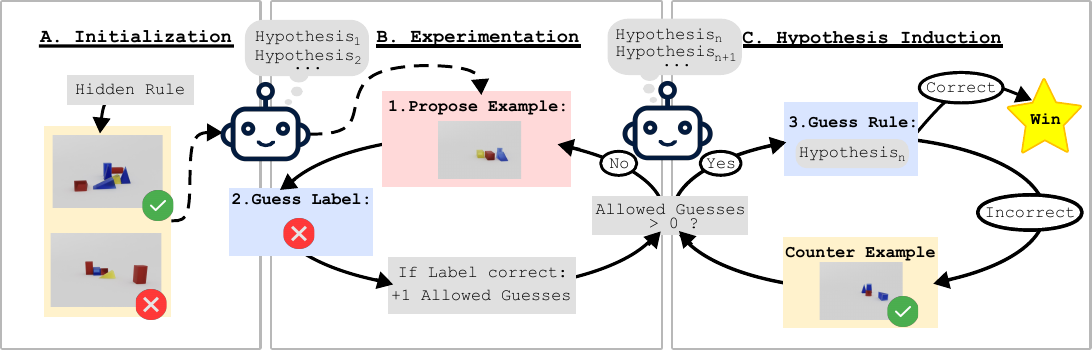}
\caption{Overview of the \environment setup. (\textbf{A}) The agent is initialized with a set of labeled seed scenes that are consistent with a hidden rule. (\textbf{B}) In the experimentation phase, the agent proposes new scenes, predicts their labels, and earns additional guess attempts for each correct prediction. (\textbf{C}) When the agent decides to commit, it submits a rule hypothesis. If incorrect, a counterexample is revealed and the agent re-enters the experimentation phase. The episode ends when the agent guesses the rule correctly or exhausts its example budget. Actions where the agent has to perceive inputs are yellow, actions where the agent needs to perform logical induction are blue and actions where experimentation is performed are red.}
\label{fig:overview}
\end{figure}

\textbf{Visual Experimentation.} A complementary line studies agents acting in visual environments, but typically under reward-driven objectives. Benchmarks such as Atari~\citep{waytowich2024atari}, NetHack~\citep{kuttler2020nethack}, and MineDojo~\citep{fan2022minedojo} evaluate VLM-based policies that map observations to actions, often via prompting or hybrid RL pipelines. Visual Agentic AI~\citep{marsili2025visual} extends this to programmatic spatial reasoning via APIs. These approaches demonstrate strong perceptual control, but focus on implicit policy optimization rather than explicit rule recovery. CLEVR-AVR~\citep{Zhou2025PhysVLMAVRAV} moves closer to active reasoning in visual domains, but agents intervene on scenes without an explicit hypothesis space or rule representation.

\textbf{Induction through Experimentation.} A third line combines induction and experimentation, primarily in symbolic domains. Zendo-based setups~\citep{bramley2018grounding} require agents to propose examples to discriminate between rules. \cite{piriyakulkij2024doing} introduce LLM-SMC-S and evaluate it on symbolic Zendo and ActiveACRE, which integrates LLM-based proposal with sequential Monte Carlo inference and builds on ACRE and the classic Blicket paradigm for studying causal concept induction \citep{Gopnik2000DetectingBH, zhang2021acre}. Related benchmarks such as AutumnBench~\citep{warrier2025benchmarking} and ARC-AGI-3~\citep{arcagi3_2026} study active discovery over symbolic grids, while Science-Gym~\citep{cerrato2026science} explores scientific discovery in a predefined parameter space with simple 2D scenes.

Thus, existing benchmarks leave grounded 3D visual rule discovery through active experimentation largely unstudied.

\section{\environment}
We propose \textbf{\environment}, an interactive reasoning benchmark inspired by the inductive logic game \emph{Zendo}, that integrates all three components within a single evaluation loop. The environment provides precise ground-truth feedback and a formal DSL-defined rule space, enabling clean separation of perception, induction, and experimentation as distinct contributors to success or failure (\cf \autoref{fig:overview}). This design supports direct comparison across method classes, including end-to-end VLMs, symbolic inference approaches, and program synthesis pipelines.
\autoref{fig:overview} provides an overview of the game. Briefly, an agent observes labeled visual scenes, infers a rule based on them and proposes new scenes to test and revise its hypotheses.

\subsection{Game Setup}

\begin{figure}[t]
    \centering
    \includegraphics[width=0.99\linewidth]{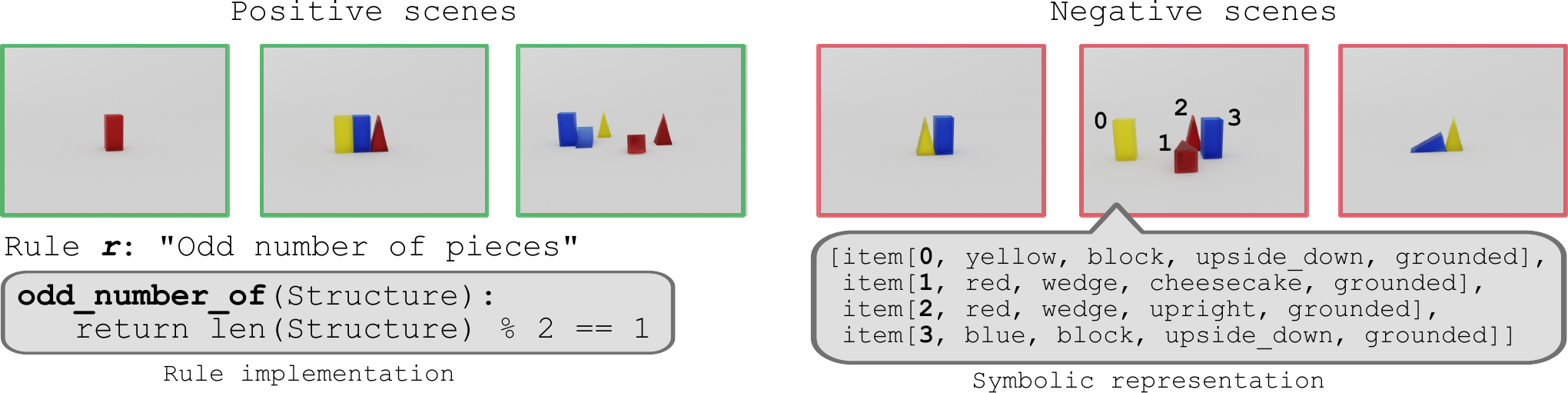}
    \caption{Scene examples for the rule "\textit{odd number of pieces}". Each scene has both a visual representation (image) and a symbolic representation.}
    \label{fig:rule_example}
\end{figure}

A \environment game episode is defined by a latent ground-truth rule \(r^*\), drawn from a rule space $\mathcal
R$ expressed in a Prolog-based DSL. The episode unfolds over a number of turns \(t \in \{1, \dots\}\), with \(T \in \mathbb{N}\). Throughout the episode, the agent accumulates a set of labeled observations \(\mathcal{D} = \{(x_i, y_i)\}\), each consisting of a visual scene \(x_i \in \mathbb{R}^{H \times W \times 3}\) and a binary label \(y_i \in \{0, 1\}\), where \(y_i = 1\) if the scene satisfies \(r^*\) and \(y_i = 0\) otherwise. The episode is bounded by the amount of total examples accumulated over the course of the game \(|\mathcal{D}| \le 30\). \autoref{fig:rule_example} illustrates example observations for the rule \emph{odd number of pieces}.

\textbf{Scene Representations.}
Each image \(x_i\) corresponds to an underlying symbolic state \(s_i \in \mathcal{S}\) that explicitly encodes object attributes (e.g., color, shape, orientation) and their spatial arrangements (e.g., \emph{touching}, \emph{stacked}, or \emph{pointing at}). The latent rule is formally defined as a Boolean classifier over the symbolic space, \(r^*: \mathcal{S} \rightarrow \{0, 1\}\), so that labels satisfy \(y_i = r^*(\Phi(x_i))\), where \(\Phi: \mathbb{R}^{H \times W \times 3} \rightarrow \mathcal{S}\) maps an image to its symbolic state \(s_i = \Phi(x_i)\). The agent must therefore bridge the high-dimensional visual input and the logical structure of \(\mathcal{S}\) to perform induction.

\textbf{Agent Interface.}
At each turn, the agent observes \(\mathcal{D}\) as (image, label) pairs. Proposed scenes \(e_t\) are emitted as structured specifications and rendered by the environment (\autoref{sec:image_gen}); hypotheses \(h_t\) may be DSL programs or natural language. This decouples perception from induction on the input side while keeping the action space symbolic and verifiable. 

\textbf{The Interaction Loop.}
The game proceeds in three phases, shown in \autoref{fig:overview}:

\begin{enumerate}[label=\textbf{\Alph*.}]
    \item \textbf{Initialization:} The agent receives two seed examples, \((x_{\text{pos}}, 1)\) and \((x_{\text{neg}}, 0)\), ensuring it starts with one positive and one negative example.
    \item \textbf{Experimentation:} At turn \(t\), the agent proposes an experiment by generating a new scene \(e_t \in \mathcal{S}\) together with a predicted label \(\hat{y}_t \in \{0, 1\}\). The environment then returns the ground-truth feedback \(y_t = r^*(e_t)\).
    \item \textbf{Hypothesis Induction:} The agent may propose a candidate rule \(h_t\) only if its prediction matched the feedback (\(\hat{y}_t = y_t\)). If \(h_t\) is logically equivalent to \(r^*\), the episode is resolved successfully. Otherwise, a counterexample \(c_{t} \in \mathbb{R}^{H \times W \times 3}\) is provided, a scene where \(h_t(\Phi(c_{t})) \neq r^*(\Phi(c_{t}))\), and the next turn begins again at Experimentation.
\end{enumerate}

After each turn, the new observations are appended to \(\mathcal{D}\). The episode terminates either when \(|\mathcal{D}| > 30\) is reached or when a proposed hypothesis \(h_t\) is logically equivalent to \(r^*\).

Equivalence between \(h_t\) and \(r^*\) is checked by canonicalizing both DSL programs and testing syntactic identity; natural-language hypotheses are first translated into the DSL by an LLM. If translation fails, an LLM judge adjudicates. As this fallback is the least reliable step, we report its invocation rate per agent in \autoref{pg:rule_guessing}.  

\paragraph{Image Generation.}
\label{sec:image_gen}
Scenes are generated via a procedural Blender–Prolog pipeline. Each scene consists of one to seven colored geometric objects (blocks, wedges, pyramids) with associated symbolic descriptions, yielding a combinatorial space of approximately \(10^{18}\) configurations.

All images are produced via rejection sampling of image descriptions. Given a rule \(r^*\) and target label \(y\), scenes are sampled uniformly and retained if \(r^*(\Phi(e)) = y\). This procedure governs both:
(i) \textbf{Seed examples}, sampled conditioned on positive/negative labels, and 
(ii) \textbf{Counterexamples}, sampled from the discriminating set \(\{s \in \mathcal{S} : h_t(s) \neq r^*(s)\}\).

As a result, feedback consists of unbiased witnesses rather than optimized or adversarial examples, ensuring consistency across the interaction.

\paragraph{Rule Space.}
\label{sec:rule_space}
The rule space \(\mathcal{R}\) is generated by a grammar \(\mathcal{G}\) inspired by the Looney Labs Zendo card set\footnote{\url{https://www.looneylabs.com/games/zendo}}, comprising predicates over object attributes (color, shape, orientation), spatial relations (e.g., \texttt{touching}, \texttt{stacked}, \texttt{pointing at}), and quantifiers. The complete grammar specification is provided in \autoref{app:grammar}. Each rule is compiled into a program in our Prolog-based DSL, which serves as the canonical representation for equivalence checking and label evaluation. The rules used in our evaluation are listed in \autoref{app:tab:tasks}.

Rules are partitioned into three categories of increasing structural complexity. \textbf{First-Order} rules apply counting predicates, such as \texttt{exactly} or \texttt{at\_least}, to individual attributes (e.g., \emph{exactly one blue upright piece}). \textbf{Second-Order} rules use comparative or parity-based quantifiers, such as \texttt{even}, \texttt{odd}, \texttt{same\_amount}, or \texttt{more\_than} (e.g., \emph{more red pieces than blue pieces}). \textbf{Complex} rules combine these constructs with spatial relations or logical connectives (e.g., \emph{a pyramid touching an upright piece}). This taxonomy enables a stratified analysis of agent performance across rules of increasing structural complexity.

\section{Experimental Evaluation}\label{sec:experiment}
We structure our evaluation around four research questions:  
\begin{enumerate}[label=Q\arabic*]
    \item How well do the agents solve \environment games end-to-end?
    \item How much does perception limit performance?
    \item How well do agents perform rule induction, particularly as complexity increases?
    \item How effective is agent-driven experimentation at reducing the hypothesis space?
\end{enumerate}

\subsection{Evaluated Agents}

We evaluated four visual rule-learning agents on challenging \environment games, which span neural, probabilistic, and neuro-symbolic approaches. The agents differ primarily in their degree of structure, access to a pretrained vision model or general VLM, use of a DSL, and experimentation strategy (\autoref{tab:agent_comparison}; details in \autoref{app:Agents}, prompts in \autoref{appendix:prompts}).

\paragraph{Oracle Agent (Neuro-Symbolic Upper Bound).}
The \textit{Oracle Agent} serves as an approximate upper bound (\autoref{tab:agent_comparison}, row 1). It combines pretrained perception with full DSL access and performs program synthesis using DeepSynth-style search \citep{fijalkow2022scaling}, paired with heuristic experiment selection that targets maximal disambiguation between candidate rules. We additionally consider an \textit{Oracle Agent (Random)} that replaces this strategy with random experiments, isolating the role of experimentation.

\paragraph{VLM Agent (End-to-End Neural Baseline).}
The \textit{VLM Agent} is the least structured approach (\autoref{tab:agent_comparison}, row 3), operating end-to-end without DSL access or explicit hypothesis search. Rule induction and experiment selection are both handled implicitly through prompting the vision-language model, providing a flexible but weakly constrained baseline.

\paragraph{Bayesian Agent (LLM-SMC Hypothesis Inference).}
The \textit{Bayesian Agent} introduces structure without requiring a DSL (\autoref{tab:agent_comparison}, row 4). It uses an LLM to propose hypotheses and maintains a particle-based posterior via sequential Monte Carlo \citep{piriyakulkij2024doing}, which also guides experiment selection. We extend prior work by incorporating image-based prompting and allowing explicit rule proposals.

\paragraph{Vision-Language Programs Agent (Program Synthesis with Learned Concepts).}
The \textit{Vision-Language Programs Agent} combines VLM-based concept extraction with program search over a general DSL \citep{wust2025synthesizing} (\autoref{tab:agent_comparison}, row 5). Unlike the Oracle, it does not assume predefined symbolic attributes, instead learning them from visual input. We extend this approach to the interactive setting by enabling VLM-guided experimentation.

For the symbolic-input ablation, the prompts (\autoref{appendix:prompts}) were adapted by replacing images with Prolog-style scene descriptions (\autoref{app:dataset_details}). 

\begin{table}[t]
\centering\small
\caption{\textbf{Comparison of agent architectures.} Type describes the main architecture of the agents; DSL refers to whether agents use any DSL; Prior refers to prior knowledge over the scenes, e.g., whether agents use a vision model trained on the scenes; Search refers to how the hypothesis space is searched; Experimentation refers to the experimentation method used when proposing examples. Agents with Zendo-specific prior knowledge are highlighted in grey.}
\label{tab:agent_comparison}
\begin{tabular}{lccccc}
\toprule
\textbf{Agent} & \textbf{Type} & \textbf{DSL} & \textbf{Prior} & \textbf{Search} & \textbf{Experimentation} \\
\midrule
\rowcolor{gray!15} Oracle        & Neuro-sym. & Full    & High & Program Synthesis & Heuristic \\
\rowcolor{gray!15} Random Oracle & Neuro-sym. & Full    & High & Program Synthesis & Random \\
VLM           & Neural     & None    & VLM  & Implicit            & Prompted \\
Bayesian      & LLM+SMC    & None    & VLM  & SMC (posterior)     & Posterior-guided \\
VLP           & Neuro-sym. & General & VLM  & Program Synthesis   & Prompted + rules \\
\bottomrule
\end{tabular}
\end{table}

\subsubsection{Human Study}
\label{sec:human_study}
To place agent performance in the context of human capability, we collected behavioral data from 19 participants playing via a browser-based implementation of \environment deployed on a JATOS server \citep{lange2015just}, totaling 10 data points per game. Participants completed an interactive tutorial and then played one to six games, which are a subset of the original evaluation set of games, following the same interaction protocol as the agents, with a maximum of 30 observations per episode. Full details of the platform, interaction protocol, and data collection are provided in \autoref{app:human_study}. The results are stated in combination with the agent results.

\subsection{Experimental Setup}

\textbf{Data.} We consider a set of $28$ games of \environment with different hidden rules, 9 rules per rule class (First-Order, Second-Order and Complex) plus one out-of-distribution (OOD) rule that cannot be expressed in the DSL of the \textit{Oracle Agent} (\cf \autoref{app:tab:tasks}). All agents are evaluated on each game across five seeds, following a common episode protocol with a maximum of \(|\mathcal{D}|=30\) observations per game.

\textbf{Models.} For the \textit{Oracle Agent} we use a pretrained vision model specified in \autoref{sec:Model}. For the other agents that rely on visual input, we report the results using 4 different VLM models: \texttt{\footnotesize gpt-5-mini-2025-08-07, gpt-5.6-terra, gemma-4-31B-it, Qwen3.8-27B} referred to as Mini, Terra, Gemma and Qwen. For the GPT variants we use low reasoning and for the open-weight models we use the recommended default values.

\textbf{Metrics.} To answer Q1 and Q2 we report the average win rate for the agents. A game is won if the final hypothesis is logically equivalent to the hidden rule. We additionally report scene-classification accuracy, computed over all classification guesses made during gameplay.
For Q3 we additionally investigate intermediate hypotheses produced by each agent. We represent each hypothesis as a symbolic program and compare it to the ground-truth rule using a structural F1 score over the rule tree. This metric measures overlap between predicates and logical operators under a permutation-invariant matching of commutative subexpressions. For each intermediate guess, we additionally measure how many of the seen examples they label correctly.
Finally, for Q4 we use an \textit{expected information gain} (EIG) metric that measures, relative to the current hypothesis posterior, how much uncertainty over possible rules a given example would resolve. The full EIG definition and experimental details are given in \autoref{app:eig}.

\subsection{Experimental Results}

\begin{figure}[t]
  \begin{minipage}[t]{0.52\linewidth}
    \vspace{7pt}
    \centering\small
    \captionsetup{type=table}
       \caption{\textbf{Agent evaluation results on \environment.} Values show mean $\pm$ SD across seeds: win rate (\%) on top and label accuracy (\%) below. The \textbf{best} and \underline{second-best} means across all agent--backbone combinations are highlighted separately for each metric. Agents are evaluated on all 28 tasks; humans on the six-task study subset, with SD across participants.}
        \label{tab:results_combined}
    \renewcommand{\arraystretch}{1.2}

\input{figures/results_table}
  \end{minipage}
  \hfill
  \begin{minipage}[t]{0.45\linewidth}
    \vspace{0pt}
    \centering
    \includegraphics[width=\linewidth]{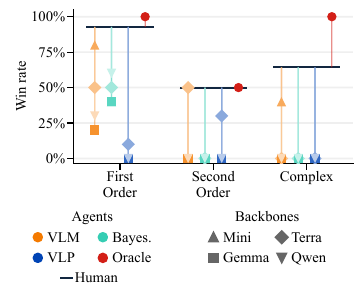}
    \captionof{figure}{\textbf{Win rate per complexity class.} Data is constrained to Human task subset, Agent results are reported as shapes, human results are reported as black line. Humans excel over VLM-based Agents, the gap becomes especially large for more challenging tasks. Different VLM backbone models are reported as different shapes.}
    \label{fig:barplot}
  \end{minipage}
  \vspace{-6pt}
\end{figure}

\textbf{\environment games are challenging for agents (Q1).}
We evaluate current AI agents on \environment to assess their ability to solve the full perception–induction loop in \autoref{tab:results_combined}, comparing against human performance on a subset of games. The total compute times for this experiment are reported in \autoref{app:compute}.

The \textit{Oracle Agent} solves 92.1\% of games. This strong performance is expected given its pretrained perception module and access to the ground-truth DSL. It fails all runs on two of the tasks and one run on the third. The failed tasks include a rule that lies outside its hypothesis space, illustrating its dependence on the true rule being expressible in the DSL.

Human participants achieve the highest win rate after the \textit{Oracle Agent} but fall behind in terms of labeling accuracy (53.9\%) in the attempted subset. Across backbone-models the performance varies particularly strongly for the \textit{VLM Agent} and \textit{VLP Agent} spanning win-rates from 7.9\% to 35.7\% and 10.7\% to 35.7\% whereas performance is more stable for the \textit{Bayesian Agent}.
Overall, \textbf{the tested methods do not robustly solve the full task}, despite strong performance on individual components. The \textit{Bayesian Agent}, for instance, achieves the highest label accuracy during interaction (80.7\%), consistent with prior work~\citep{piriyakulkij2024doing}, yet this does not translate into end-to-end success. This supports recent findings in the work of \cite{beger2025ai} that correct classification alone does not entail recovery of the underlying rules.

\autoref{fig:barplot} shows the win rates across three rule types limited to the same 6 tasks that human participants attempted to solve for a fair comparison. The second-order class includes the \textit{Oracle Agents} OOD rule which leads to the decline in this class. Overall the performance of the learned agents declines with complexity of the task across backbone-models whereas Human performance remains above 50\% even on complex tasks.
The \textit{Bayesian Agent} succeeds only on first-order logic games and fails entirely on more complex settings in this set of tasks. In contrast, the \textit{VLP Agent} maintains relatively consistent performance across first-order and second-order games. The \textit{VLM Agent} performs best across the learned agents in this subset but no single backbone-model solves all classes. Human participants achieve at least a 50\% win rate in all game types. This indicates a \textbf{gap between VLM-based rule induction and human-like rule learning,} particularly as rule complexity increases.

\textbf{Perception is a bottleneck for inducing correct rules (Q2).}
To disentangle perception from induction, we evaluate all agents in a symbolic-input setting, replacing rendered scenes with ground-truth Prolog-style descriptions (\autoref{tab:results_symbolic}).

Symbolic input improves all four \textit{VLP} configurations and three of four \textit{VLM} configurations, whereas \textit{Bayesian} performance improves only with Mini, confirming that perception is a key bottleneck. The \textit{VLP Agent} benefits most, gaining over $45\%$ with Gemma and solving more than half of the games. The \textit{VLM Agent} shows a strong improvement with Terra, gaining $25\%$. The \textit{Bayesian Agent} improves only using Mini but remains the weakest overall, solving fewer than one third of the games. Part of the perception bottleneck may stem from the mismatch between Blender-generated scenes and the natural images used during pretraining, limiting performance on generated scenes.

\begin{wraptable}{r}{0.48\textwidth}
  \centering\small
  \setlength{\tabcolsep}{3pt}
  \vspace{-\baselineskip}
  \caption{\textbf{Agent evaluation results on Symbolic \environment.} Means across seeds $\pm$ SD across seeds. Symbolic models and their perception-based baselines use matching completed task/seed pairs within each agent. Green/red annotations show absolute percentage point change vs.\ the perception-based variant. \textbf{Best} and \underline{second best} win rates highlighted.}
  \label{tab:results_symbolic}
  \begin{tabular}{llc}
    \toprule
    \textbf{Agent} & \textbf{Model} & \textbf{Wins (\%)} \\
    \midrule
    VLM Agent & Mini & 43.6 {\tiny $\pm$ 5.3}\,\imp{+7.9} \\
    Bayesian Agent & Mini & 22.9 {\tiny $\pm$ 3.2}\,\imp{+8.6} \\
    VLP Agent & Mini & 49.3 {\tiny $\pm$ 5.3}\,\imp{+31.4} \\
    \midrule
    VLM Agent & Terra & \underline{58.6} {\tiny $\pm$ 4.1}\,\imp{+25.0} \\
    Bayesian Agent & Terra & 22.9 {\tiny $\pm$ 4.8}\,\impneg{-4.3} \\
    VLP Agent & Terra & 54.3 {\tiny $\pm$ 6.4}\,\imp{+18.6} \\
    \midrule
    VLM Agent & Gemma & 33.6 {\tiny $\pm$ 5.4}\,\imp{+25.7} \\
    Bayesian Agent & Gemma & 13.6 {\tiny $\pm$ 1.6}\,\impneg{-2.9} \\
    VLP Agent & Gemma & 56.4 {\tiny $\pm$ 3.0}\,\imp{+45.7} \\
    \midrule
    VLM Agent & Qwen & 2.1 {\tiny $\pm$ 3.2}\,\impneg{-6.4} \\
    Bayesian Agent & Qwen & 10.7 {\tiny $\pm$ 3.6}\,\impneg{-5.0} \\
    VLP Agent & Qwen & 49.3 {\tiny $\pm$ 3.9}\,\imp{+37.9} \\
    \midrule
    Oracle Agent & -- & \textbf{96.4} {\tiny $\pm$ 0.0}\,\imp{+4.3} \\
    \bottomrule
  \end{tabular}
  \vspace{-0.3cm}
\end{wraptable}

Despite these gains, no agent approaches Oracle-level performance, indicating that induction remains a fundamental bottleneck even when perception is removed. A breakdown by rule category (Appendix: \autoref{app:tab:symbolic_tasks}) shows that VLM-based agents struggle particularly with second-order and compositional rules.
Collectively, these results suggest that \textbf{both perception and induction constrain performance}, with their relative impact varying across methods.

\begin{figure}[t]
  \centering
  \includegraphics[width=1\linewidth]{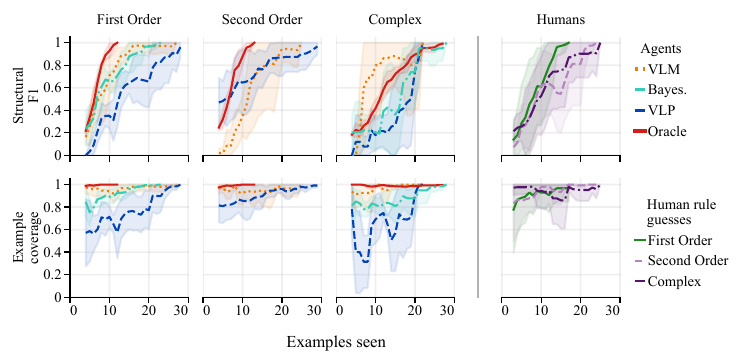}
  \caption{\textbf{Rule-guess quality versus number of observed examples.} Left: Agent games across 28 tasks, grouped by rule type, with five seeds using Terra as the backbone model. Right: Human games on the study subset, grouped by rule type. Top: Structural F1 in successful games; the latest guess is carried forward up to the last recorded guess within each agent-rule-type group. Bottom: Coverage of intermediate rule guesses in won games on seen examples. Shading indicates $\pm1$ SD across participant-level task means for humans and across seed-level task means for agents.}
  \label{fig:rule_eval}
  \vspace{-10pt}
\end{figure}

\textbf{Rule Induction is fastest for first-order rules and slows down on more complex rules (Q3).}
To better understand the space of hypotheses the agents explore, we compute a structural F1 curve per game as a function of the number of observed examples, and then average these curves across games to obtain the aggregate learning trajectory. \autoref{fig:rule_eval} shows the structural F1 score of the intermediate rule guesses as a function of the number of observed examples from the evaluations using \texttt{\footnotesize gpt-5.6-terra} due to this backbone model leading to the highest solve rates on average.

The top row examines hypothesis refinement conditional on episode success. Mean structural F1 generally increases with additional examples, indicating convergence toward the target rule. \textbf{Overall, first-order rules are refined more quickly}: most agents reach high structural similarity, and a perfect score of 1, with fewer examples than on second-order or complex tasks. The \textit{VLP Agent} is an exception, showing strong refinement on second-order tasks, possibly because short parity programs are readily accessible in its DSL and search procedure. Human refinement resembles the \textit{VLM Agent} on first-order tasks and the \textit{Oracle Agent} on second-order and complex tasks. Within the harder rule classes, however, the ordering is not consistent across players; for example, human participants reach high structural similarity later on second-order than on complex tasks.

The bottom row of \autoref{fig:rule_eval} measures how consistently guessed rules classify the labeled examples available at the time of each guess, restricting the analysis to won games. The \textit{Oracle Agent} achieves the highest observed-example coverage, demonstrating that hypotheses can remain consistent with much of the available evidence. In contrast, the \textit{VLP Agent} often proposes rules that correctly classify only slightly more than half of the observed scenes. This may reflect errors in backbone-derived scene representations or concepts propagating into program search, although limitations of the search procedure or hypothesis space could also contribute. The \textit{Bayesian Agent} and \textit{VLM Agent} achieve similar accuracy, but neither consistently explains all observations, even though the \textit{Bayesian Agent} computes predicted labels for each hypothesis and selects the best-performing one. The example coverage of human guesses is most similar to the \textit{VLM Agent} hypotheses. Human participants expressed difficulty finding hypotheses in later stages that cover all examples, which could explain the dip in performance in second-order and complex games at around 15 examples.

\textbf{Agent-driven experimentation does not consistently outperform random scene generation in informativeness (Q4).}
To assess how proposed experiments contribute to convergence, we compare their expected information gain (EIG), including a \textit{Random Oracle Agent} variant that generates random scenes. We use the data collected during the symbolic \environment experiment with \texttt{\footnotesize gpt-5.6-terra} as the shared backbone model.

\begin{wrapfigure}{r}{0.4\textwidth}
  \centering
  \vspace{-20pt}
  \includegraphics[width=1\linewidth]{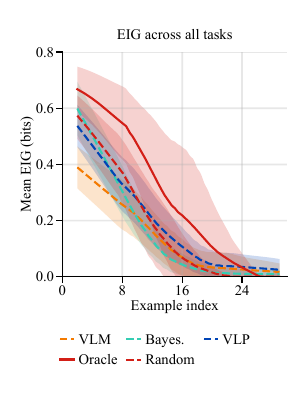}
  \caption{\textbf{Expected information gain (EIG, bits) by example position within the game}, averaged across tasks and seeds. Shaded bands show $\pm1$ SD across seeds. Data from symbolic \mbox{\environment} experiments.}
  \label{fig:eig}
  \vspace{-20pt}
\end{wrapfigure}

As shown in \autoref{fig:eig}, the \textit{Oracle Agent} achieves the highest mean EIG (0.542), while random scene generation performs second best (0.392). VLM-based agents generally achieve lower EIG values. Among them, the \textit{VLP Agent} performs slightly better, with a mean EIG value of 0.245, whereas the \textit{Bayesian Agent} and \textit{VLM Agent} remain below 0.2. All Agents show a decline in EIG scores as more examples are observed.

Despite selecting among ten candidate experiments using its hypothesis distribution, the \textit{Bayesian Agent} does not consistently outperform direct VLM-based experiment generation. This suggests that hypothesis-guided selection is insufficient to guarantee more informative experiments; its effectiveness likely depends on both the diversity and quality of the candidate scenes and the hypotheses used to evaluate them. The comparatively low EIG of the VLM-based agents may also be related to the high rate of repeated or recycled examples discussed in \autoref{app:recycling}.

Overall, random scene generation provides a surprisingly strong baseline, motivating further investigation into how agents can combine diverse exploration with effective hypothesis-guided experiment selection.

\paragraph{Discussion \& Future Work.}
The results reveal distinct failure modes across the evaluated agents. The \textit{Oracle Agent} succeeds via its tailored perception module and DSL access, but fails when the hidden rule requires an out-of-vocabulary predicate. Among VLM-based methods, the \textit{VLM- and VLP Agent} perform best end-to-end yet still fail on over half of episodes, especially on relational and compositional rules. The \textit{Bayesian Agent} excels at labeling but only recovers short, simple rules; merging hypotheses via logical connectives and providing more observations in the prompt could push it toward more complex ones. The \textit{VLP Agent} handles predicates like \texttt{EVEN/ODD}, but structured search alone is insufficient when the predicate inventory is incomplete or the generalized DSL inflates the search space. Overall, \environment exposes a clear gap between solving individual subcomponents and the full interactive loop.

A deeper architectural question concerns the strict separation between perception and reasoning itself. The \textit{Oracle Agent}'s modular design is interpretable and efficient, but its perceptual vocabulary is fixed before rule learning begins: the DSL predicates define not only what can be reasoned about, but also what can be \emph{seen}.
\cite{hofstadter1995fluid} argues that human conceptual thinking does not proceed with such clean separation; perception and abstract reasoning are mutually constitutive, with active concepts reshaping attended features and novel groupings giving rise to new distinctions. A more principled Zendo player might therefore embed perception within the hypothesis-testing loop, letting the current best hypothesis direct visual attention toward features most relevant for discriminating among remaining candidates.

\section{Conclusion}
We introduced the benchmark \environment, a challenging, controlled interactive environment that integrates perception, induction, and experiment design within a single closed loop, and used it to investigate the core challenges of active visual rule learning. Our results establish three main findings. First, strong performance on example labeling does not guarantee correct hypothesis induction, with important implications for how inductive reasoning systems should be evaluated. Second, perception and induction failures are distinct and agent-dependent: removing visual input substantially improves the \textit{VLP- and VLM Agent}, but has little effect on the \textit{Bayesian Agent}, suggesting that the primary bottleneck lies in induction and hypothesis revision rather than perception. Third, the VLM-based agents consistently struggle to actively reduce hypothesis uncertainty, showing low information gain under our hypothesis model, even when perception is removed. This persistent failure points to a structural limitation in how these agents explore and search the hypothesis space.

These findings suggest that progress in active rule learning will require architectures that combine visual grounding with structured, falsifiable hypotheses and active search. Two promising directions are: (1) tighter feedback loops between hypotheses and observations, allowing agents to revisit examples and resolve inconsistencies, and (2) hybrid symbolic-neural systems where structured hypotheses guide perceptual feature extraction instead of relying on a fixed perception module.

\section*{Acknowledgments}
This project was partially funded by the Deutsche Forschungsgemeinschaft (DFG, German Research Foundation) under Germany's Excellence Strategy (EXC-3057/1 „Reasonable Artificial Intelligence“, Project No. 533677015).
This work was further supported by the Priority Program (SPP) 2422 in the subproject “Optimization of active surface design of high-speed progressive tools using machine and deep learning algorithms“ funded by the German Research Foundation (DFG). %
It has also benefited from the Cluster of Excellence "The Adaptive Mind", funded by the DFG under Germany's Excellence Strategy - EXC-3066. This work was further funded in part by the Deutsche Forschungsgemeinschaft (DFG, German Research Foundation) – GRK 2853 “Neuroexplicit Models of Language, Vision, and Action” (Project No. 471607914).

Kevin Ellis and Wasu Top Piriyakulkij acknowledge support from an NSF CAREER grant.

The authors acknowledge the contributions of Christian Kasiman, Matthias Petri, Kilian Schwarz, Ronik Tempski and Paul Jakob Weitz to an initial student project, whose code served as a starting point for the extended implementation presented in this paper.

\bibliographystyle{plainnat}
\bibliography{references}  

\appendix
\include{appendix}

\end{document}

%% file: figures/results_table.tex
\begingroup
  \centering
  \small
  \setlength{\tabcolsep}{3pt}
  \renewcommand{\arraystretch}{1.1}
  \newcommand{\SD}[1]{{\tiny$\pm$#1}}
\begin{adjustbox}{max width=0.95\linewidth}
\begin{tabular}{ccccc|c}
  \toprule
  \multicolumn{6}{c}{\textbf{Win rate (\%) $\uparrow$}} \\
  \midrule
    & \textbf{VLM} & \textbf{Bayesian} & \textbf{VLP}
    & \textbf{Oracle} & \textbf{Human} \\
  \midrule
  Mini  & \underline{35.7}\SD{13.1} & 14.3\SD{3.6} & 17.9\SD{2.5}
        & \multirow{4}{*}{\textbf{92.1}\SD{1.6}}
        & \multirow{4}{*}{73.3\SD{36.9}} \\
  Terra & 33.6\SD{2.0} & 27.1\SD{4.1} & \underline{35.7}\SD{3.6} & & \\
  Gemma & 7.9\SD{3.0} & 16.4\SD{4.1} & 10.7\SD{0.0} & & \\
  Qwen  & 8.6\SD{2.0} & 15.7\SD{7.0} & 11.4\SD{1.6} & & \\
  \midrule
  \multicolumn{6}{c}{\textbf{Accuracy (\%) $\uparrow$}} \\
  \midrule
    & \textbf{VLM} & \textbf{Bayesian} & \textbf{VLP}
    & \textbf{Oracle} & \textbf{Human} \\
  \midrule
  Mini  & 52.5\SD{4.4} & 76.0\SD{3.9} & 66.0\SD{1.5}
        & \multirow{4}{*}{70.8\SD{2.9}}
        & \multirow{4}{*}{53.9\SD{14.3}} \\
  Terra & 36.2\SD{1.4} & \underline{79.6}\SD{1.7} & 62.2\SD{2.1} & & \\
  Gemma & 45.3\SD{1.7} & 78.2\SD{1.7} & 59.2\SD{1.4} & & \\
  Qwen  & 48.4\SD{4.8} & \textbf{80.7}\SD{1.9} & 72.1\SD{5.0} & & \\
  \bottomrule
  \end{tabular}
  \end{adjustbox}
  \par
\endgroup

%% file: appendix.tex
\section{Agents}
\label{app:Agents}
\subsection*{Oracle Agent}
The Oracle Agent perceives each Zendo scene by passing its rendered image through a pre-trained Vision-Model, a transformer-based object encoder that produces a 7×15 integer tensor, encoding each piece's object ID, color, shape, orientation, six touching-relation slots, and pointing target. To propose a new experiment, the agent first runs a bigram-prior PCFG heap search over the hand-crafted Zendo DSL which is a typed library of predicates defines in \autoref{app:dsl_details} leaving out the binary rule \texttt{SAME\_AMOUNT}. The search enumerates programs in descending prior-probability order while filtering known-incorrect and canonically equivalent programs. The bigram-prior PCFG is a probabilistic context-free grammar over the DSL whose production probabilities are conditioned on the parent rule and argument index, with the conditional weights being uniformly distributed. If no fully consistent program is found, the accuracy threshold is relaxed by one example at a time until a candidate survives. The top DSL program is then compiled back to Prolog to function calls. The primary proposal strategy calls the Prolog engine with a query including both Prolog function calls, generating a scene that either satisfies $rule_2$ but not $rule_1$ or the other way around. If only one candidate exists, it falls back to generate an example that either aligns with the rule or not. The resulting Prolog scene string is rendered to a PNG via Blender. To label an externally proposed scene, the agent evaluates the top-ranked DSL program on its tensor encoding. To guess the rule, it returns the highest-scoring program from the heap search.

\subsection*{VLM Agent}
The VLM Agent delegates all perception, generation, and inference entirely to a multimodal large language model, with no symbolic program search. Each observed example is stored as a raw (tensor, label) pair alongside its rendered image path, the tensor is computed once from the Prolog scene string but is not used for reasoning. To propose a new experiment, the agent encodes all positive and negative example images as base64 data URIs, constructs a multimodal prompt interleaving these images with a task description, and instructs the LLM to output a Python literal of the form [[item(id, color, shape, orientation, grounding), ...], label]. The proposed label is cached and returned verbatim when the game engine subsequently asks the agent to classify the scene it just proposed, avoiding a redundant LLM call. To guess the rule, the agent assembles an analogous multimodal prompt from all labeled example images and requests a natural-language rule. The response is returned as-is without compilation or consistency verification against the examples.

\subsection*{Bayesian Agent}
The Bayesian Agent maintains a particle approximation to a posterior over natural-language hypotheses, combining LLM-based hypothesis generation with probabilistic filtering over concept descriptions. Each hypothesis is a free-form English rule (e.g., ``there must be a red block touching a pyramid''), which is evaluated by first translating it into executable code using the rule-conversion prompt listed in \autoref{appendix:prompts}. The resulting program is then executed on a scene to determine whether the hypothesis predicts a positive or negative label.

On the first turn, the agent selects a random positive example and, for each of seven attribute families (colors, shapes, orientations, groundedness, touching, pointing, and stacking), prompts to generate five candidate hypotheses conditioned on that image. These candidates form the initial particle set.
Given a labeled scene, each particle is scored by its marginal likelihood under a noisy concept-learning model with recall parameter \(\theta\) and specificity parameter \(\delta\). We place Gaussian priors on \(\theta\) and \(\delta\), with \(\theta \sim \mathcal{N}(0.7, 0.1)\) and \(\delta \sim \mathcal{N}(0.9, 0.01)\), truncated to \([0.5,1.0]\). For a scene \(x\) with label \(y\), the likelihood is high when the translated hypothesis predicts \(y\) and lower otherwise, with false positives and false negatives weighted asymmetrically according to \(\theta\) and \(\delta\). The final particle weight is obtained by marginalizing this likelihood over the \((\theta,\delta)\) prior grid.

Particles are resampled to a support of 25 hypotheses using systematic resampling. After each new observation, the lowest-likelihood particles are rejuvenated by prompting the LLM with the latest labeled scene and the current hypothesis to propose refined variants; only refinements that improve the joint score are retained. To propose a new experiment, the agent estimates the expected KL divergence between the posterior before and after observing the outcome of each candidate scene, and selects the scene with the highest expected information gain. Candidate scenes are generated by prompting the LLM with the current high-weight hypotheses to produce Prolog-style scene descriptions for each label. To label a scene, the agent marginalizes \(P(\mathrm{yes}\mid x,c)\) over the particle support. To guess the rule, it returns the MAP particle, i.e., the most frequent hypothesis after resampling.

\subsection*{VLP Agent}
The Visual Language Programming Agent induces a concept vocabulary from images before searching for rules. Upon each of the first 5 observations, the agent queries a vision-language prompter with the newly received image to discover novel object categories (target: 6), visual properties (target: 9), and spatial actions/relations (target: 5) not yet in its vocabulary. Each newly identified concept string is registered as a typed DSL primitive (objects as OBJECT, properties as PROPERTY, relations as ACTION) dynamically extending the grammar. Subsequent rounds include a "Already Discovered" header so the VLM only returns genuinely new vocabulary. Program search then runs a bigram-prior PCFG heap search over this evolving VLP DSL. During evaluation a VLM is prompted using raw image paths rather than fixed tensors to detect pieces in images, letting the language model act as the semantic interpreter. To propose a new experiment, the agent runs the VLP search, takes the top two surviving programs, and constructs a multimodal prompt with all labeled example images and the candidate rules. The VLM responds with a [[item(...), ...], label] Prolog literal that is parsed, validated against the 7×15 tensor schema, and rendered via Blender. To label a scene or guess a rule, the agent returns the result of the top candidate, evaluated on the image or the string form of the top-ranked VLP program respectively.

\section{Extended Evaluations}
To assess whether the trends observed in the main experiment persist across a broader set of games, we evaluate the four agents on an additional 72 games, yielding 100 game runs in total when combined with the 28 games from the main experiment. The rules were selected such that each predicate occurs 4 times and compositional rules using \texttt{AND} and \texttt{OR} connectives occur 6 times leading to \(22 \times 4 + 12 = 100\) rules. We evaluate the VLM-based agents using \texttt{Gemma-4-31B-it} and \texttt{Qwen3.8-27B}, with one seed per backbone. The results are summarized in \autoref{tab:results_100}.

\begin{table}[ht]
  \centering
  \caption{Player results on 100 games per Agent. Label accuracy is correct quiz labels divided by all quiz labels. Average turns uses winning games only.}
  \label{tab:results_100}
  \begin{tabular}{lrrrr}
    \toprule
    Player & Model & Wins (\%) & Avg Turns & Label Acc (\%) \\
    \midrule
    VLM & Gemma & 9/100 (9.00\%) & 6.22 & 50.36 \\
    VLM & Qwen & 5/100 (5.00\%) & 4.20 & 53.28 \\
    Bayesian & Gemma & 12/100 (12.00\%) & 6.75 & 77.29 \\
    Bayesian & Qwen & 6/100 (6.00\%) & 6.33 & 77.28 \\
    VLP & Gemma & 21/100 (21.00\%) & 5.48 & 63.04 \\
    VLP & Qwen & 11/100 (11.00\%) & 2.73 & 78.95 \\
    Oracle & - & 89/100 (89.00\%) & 5.13 & 70.43 \\
    \bottomrule
  \end{tabular}
\end{table}

Bare \texttt{EVEN} and \texttt{ODD} without attributes were the easiest predicates for VLP: both backbones solved all four occurrences of each, whereas \textit{VLM Agent} and \textit{Bayesian Agent} solved none.

The rule \texttt{ALL\_THREE\_COLORS} was easiest for the Gemma \textit{VLM Agent} and \textit{Bayesian Agent}, both of which solved all four occurrences. The Qwen \textit{VLM-Agent} solved 1/4, while the Qwen \textit{Bayesian Agent} and both \textit{VLP} variants solved none.

The predicate \texttt{ZERO} with one attribute exhibited the broadest success across agent–backbone combinations. Five combinations solved 2/4 instances, with only Qwen \textit{VLP Agent} solving none. Pooled across all combinations, \texttt{ZERO} achieved the highest win rate, with 10/24 successful runs.

Performance also varied substantially across attribute variants of the same predicate. For example, both \textit{VLP} backbones solved "Even amount of blocks." and "Even amount of wedges", but neither solved the corresponding doorstop or upside-down variants. Similarly, "Zero pieces are yellow." was solved by five agent–backbone combinations, whereas "Zero pieces are blue." was solved by four. Finally, "At least 3 pieces are red." was solved by four combinations, while the corresponding cheesecake and doorstop variants were solved by none. These results suggest that difficulty is determined not only by the underlying predicate, but also by the attribute to which it is applied. However, attribute selection alone does not account for the observed failures: complex rules involving combinators were not solved by any agent–backbone combination, nor were any games whose predicates required reasoning about interactions between pieces.

\subsection{Expected Information Gain}
\label{app:eig}
To quantify the informativeness of player-proposed examples, we compute the \emph{expected information gain} (EIG) of each example relative to the current hypothesis posterior.

\begin{wrapfigure}{r}{0.4\textwidth}
  \centering
  \vspace{-10pt}
  \includegraphics[width=1\linewidth]{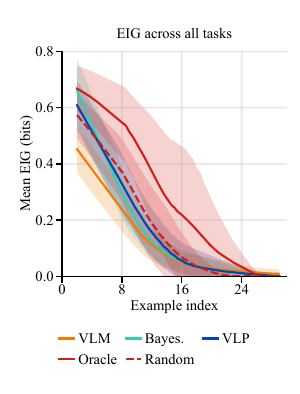}
  \caption{\textbf{Expected information gain (EIG, bits) by example position of experiment within the game}, averaged across tasks and seeds. Shaded bands show $\pm1$ SD across seeds. Data was used from symbolic \mbox{\environment} experiments.}
  \label{fig:eig_mini}
  \vspace{-10pt}
\end{wrapfigure}

Given $n$ prior labeled examples \(E = \{(x_i, y_i)\}_{i=1}^n\), we enumerate the top-\(K\) \(K=20\) most probable programs \(H = \{h_j\}_{j=1}^K \subseteq \mathcal{H}\) consistent with \(E\) using heap search~\citep{fijalkow2022scaling} over a uniform Probabilistic Context Free Grammar (PCFG) derived from the Zendo-specific DSL. Experiments for which the search finds fewer than 2 perfectly consistent programs are excluded from the EIG analysis, as the EIG then cannot be determined. This typically occurs when the examples fully specify a single rule, such that any alternative rule is inconsistent with the observed data. This happens for short rules towards the end of the game when more than 20 examples have been added.
Each \(h_j\) has a prior weight given by the PCFG \(w_j = P(h_j)\). We form a posterior over the retained hypotheses by renormalizing these weights:
\[
P(h_j \mid E) = \frac{w_j \cdot 1[h_j(\cdot) \models E]}{\sum_{h_k} w_k \cdot 1[h_k(\cdot) \models E]},
\]
ensuring that the probability mass sums to 1 over the top-\(K\) consistent hypotheses.
For a proposed example \(x^*\), the EIG is the binary entropy of the predicted label:
\[\text{EIG}(x^*) = H\!\left(\sum_{j=1}^K P(h_j \mid E) \cdot 1[h_j(x^*) = \text{True}]\right),\]
where $H(p) = -p \log_2 p - (1-p) \log_2(1-p)$. EIG $\approx 1$ when hypotheses split 50/50 (maximally informative); EIG $\approx 0$ when all agree (redundant). We report average EIG per agent as a measure of sample efficiency.

\textbf{EIG results for GPT-5-mini.}
In addition to the GPT-5.6-Terra results in the main text, we report the EIG curves per agent using GPT-5-mini in \autoref{fig:eig_mini}. Similarly to the results reported in the results section, the agents experiments on average have a lower EIG score than the random baseline.

\begin{figure}[t]
  \begin{minipage}[t]{0.45\linewidth}
    \vspace{7pt}
    \centering\small
    \captionsetup{type=table}
\caption{Repeated player-proposed examples across 28 symbolic tasks and seeds. A repeat matches the pieces and symbolic spatial relations of an earlier example. Repeat rates use all proposals as the denominator.}
\label{tab:repeated_examples}
    \begin{tabular}{ccr}
\toprule
Agent & Backbone & Repeat rate \\
\midrule
VLM      & Mini  & 9.21\%  \\
VLM      & Terra & 6.46\%  \\
Bayesian & Mini  & 27.78\%  \\
Bayesian & Terra & 25.98\% \\
VLP      & Mini  & 14.18\% \\
VLP      & Terra & 2.54\%  \\
Oracle   & --    & 0.00\% \\
Oracle (Random) & -- & 0.42\% \\
\bottomrule
\end{tabular}
  \end{minipage}
  \hfill
  \begin{minipage}[t]{0.52\linewidth}
    \vspace{0pt}
    \centering
    \includegraphics[width=1\linewidth]{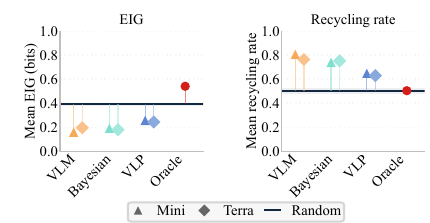}
  \caption{Left: \textbf{Mean expected information gain (EIG, bits)}, averaged across tasks and seeds. Right: \textbf{Mean rate of pieces already seen in previous examples per experiment}. Lines show $\pm1$ SD across seeds. Data was used from symbolic \environment experiments.}
    \label{fig:eig_recycling}
  \end{minipage}
  \vspace{-6pt}
\end{figure}
\subsection{Rate of recycled pieces}
\label{app:recycling}
In addition to the estimated information gain, we want to further report the rate at which experiments are repeated across turns. \autoref{tab:repeated_examples} reports how frequently agents propose experiments that exactly reproduce an example already present in the interaction history. Across agent configurations, models using the Terra backbone generally exhibit lower repeat rates than their Mini counterparts. The difference is particularly pronounced for the VLP agent, whose repeat rate decreases from 14.18\% with Mini to 2.54\% with Terra.

Beyond exact repetition, agents may also construct new experiments from pieces that have already appeared in earlier examples. \autoref{fig:eig_recycling} compares mean expected information gain (EIG) with the degree of such piece recycling. To quantify this relationship at the example level, we paired each proposed example's EIG with the fraction of its pieces that had appeared in previous examples, yielding 7,065 examples across 672 symbolic games.

A regression controlling for game fixed effects and configuration-specific example-index effects shows a negative association between piece recycling and information gain. A 10-percentage-point increase in the fraction of recycled pieces is associated with a decrease of $0.0085$ bits in EIG (95\% CI: $0.0038$--$0.0132$ bits; $p=0.00093$), with standard errors clustered by task. This association remains statistically significant within games after accounting for the position of an example in the interaction sequence, although the analysis is correlational and does not establish a causal effect of recycling on information gain.

\subsection{Intermediate Rule Guesses}
\begin{figure}[t]
  \centering
  \includegraphics[width=1\linewidth]{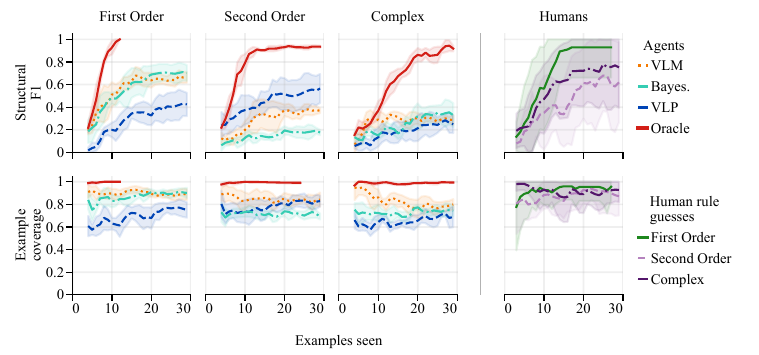}
  \caption{\textbf{Rule-guess quality versus number of observed examples.} Left: Agent games across 28 tasks, grouped by rule type, with up to five seeds per backbone and equal weighting of contributing backbones. Right: Human games on the study subset, grouped by rule type. Top: Structural F1 in all games; the latest guess is carried forward up to the last recorded guess within each agent-rule-type group. Bottom: Coverage of intermediate rule guesses in all games on seen examples. Shading indicates $\pm1$ SD across participant-level task means for humans and across seed-level task means for agents, combining seed and backbone variation with equal backbone weighting.}
  \label{fig:rule_eval_all}
  \vspace{-10pt}
\end{figure}
In addition to \autoref{fig:rule_eval} that reports the F1 scores and the sample coverage for intermediate guesses constrained to the games won, we report the results for all the games in \autoref{fig:rule_eval_all}. The total results show a gap in the ability to produce rules that correctly classify previously seen examples between human participants and learned agents. While the guesses by human participants remain on average at $90\%$ coverage, the guesses from the learned agents, specifically the \textit{VLP Agent} and \textit{Bayesian Agent}, remain below $80\%$ for most of the games.

\begin{table}[ht]
\caption{Computation of experiments across 28 tasks
and five seeds per player. Total time sums recorded elapsed turn
durations, including API waiting time; it does not measure CPU-hours.
Mean turn duration is weighted by the number of turns.}
\centering
\resizebox{\textwidth}{!}{
\begin{tabular}{lrrrrrrr}
\toprule
Player & Model & Runs & Turns & Total time (h) & Mean turn (s) & Turns/run & Tokens \\
\midrule
VLM  & Mini & 140 & 2,106 & 63.71  & 108.91 & 15.04 & 26,446,821 \\
Bayesian & Mini & 140 & 2,076 & 146.77 & 254.51 & 14.83 & 133,216,701 \\
VLP  & Mini  & 140 & 2,182 & 200.49 & 330.78 & 15.59 & 22,185,149 \\
\midrule
Combined & Mini  & 420 & 6,364 & 410.97 & 232.48 & 15.15 & 181,848,671\\
\midrule
VLM  & Terra & 140 & 2,301 & 19.64  & 30.73  & 16.44 & 17,387,258\\
Bayesian & Terra & 140 & 1,879 & 120.91 & 231.65 & 13.42 & 17,653,981\\
VLP   & Terra  & 140 & 1,976 & 320.24 & 583.44 & 14.11 & 37,130,251\\
\midrule
Combined  & Terra & 420 & 6,156 & 460.79 & 269.47 & 14.66 & 72,171,490\\
\midrule
VLM & Gemma & 140 & 3,170 & 31.63 & 35.93 & 22.64 & 14,501,902 \\
Bayesian & Gemma & 140 & 2,017 & 167.85 & 299.58 & 14.41 & 22,900,981 \\
VLP & Gemma & 140 & 2,290 & 204.38 & 321.30 & 16.36 & 26,055,460 \\
\midrule
Combined & Gemma & 420 & 7,477 & 403.86 & 194.45 & 17.80 & 63,458,343 \\
\midrule
VLM & Qwen & 140 & 2,556 & 77.33 & 108.92 & 18.26 & 18,114,448 \\
Bayesian & Qwen & 140 & 2,052 & 290.01 & 508.79 & 14.66 & 31,926,963 \\
VLP & Qwen & 140 & 2,105 & 186.52 & 319.00 & 15.04 & 17,293,941 \\
\midrule
Combined & Qwen & 420 & 6,713 & 553.86 & 297.03 & 15.98 & 67,335,352 \\
\bottomrule
\end{tabular}
}
\label{tab:nonsymbolic-timing}
\end{table} 

\section{Computational Resources}\label{app:compute}
The primary experiment comprised 28 tasks evaluated with five seeds. Three agents were evaluated with four VLM backbones each, while the \textit{Oracle Agent} was evaluated independently of a backbone, yielding 1,820 runs in total. \autoref{tab:nonsymbolic-timing} reports runtime, token usage, and mean turns per run for each agent--backbone combination. Experiments with Mini and Terra, as well as evaluations of the \textit{Oracle Agent}, used NVIDIA Tesla V100-SXM3 GPUs (32\,GB VRAM) and Intel Xeon Platinum 8174 CPUs. Experiments with Gemma and Qwen used NVIDIA A100-SXM4 GPUs (80\,GB VRAM) and AMD EPYC 7313 16-core CPUs.

These figures exclude preliminary experiments, symbolic ablations, and training of the vision model used by the \textit{Oracle Agent}.

\section{Dataset Details}\label{app:dataset_details}
\begin{wrapfigure}{r}{0.48\textwidth}
  \centering
  \vspace{10pt}
    \includegraphics[width=\linewidth]{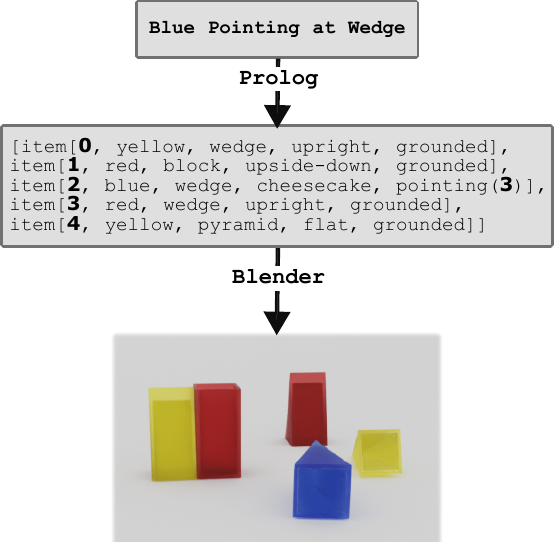}
    \caption{Generation pipeline: Rules are sampled, examples are retrieved through Prolog sampling and images are generated using Blender.}
    \label{fig:gen-pipeline}
    \vspace{-25pt}
\end{wrapfigure}

We generated a synthetic dataset of labeled Zendo scenes using Prolog and Blender shown in \autoref{fig:gen-pipeline}. Each scene depicts a structure composed of colored geometric shapes (blocks, wedges, pyramids) in different orientations, with up to seven pieces per scene. The data generation pipeline samples random logical rules from predefined templates involving logical combinators, quantifiers, attributes, and relations. Each rule is automatically translated into a Prolog query to generate examples that either satisfy or violate the rule. 

\paragraph{Object Detection Data:}
We generated 54,252 training images at $640\times480$ resolution, each annotated with per-object attributes (color, shape, orientation, bounding box) and relations (\texttt{pointing}, \texttt{touching}, \texttt{on\_top\_of}). Object encodings follow the vector format:
\begin{multline}
\label{eq:vec}
[\mathtt{ID}, \mathtt{COLOR}, \mathtt{SHAPE}, \mathtt{ORIENTATION}, \mathtt{left}, \mathtt{right}, \\
\mathtt{front}, \mathtt{back}, \mathtt{top}, \mathtt{bottom}, \mathtt{pointing}, x_{\min}, y_{\min}, x_{\max}, y_{\max}]
\end{multline}

A separate test set of 2,779 images was generated with comparable attribute distributions (e.g., $52\%$ vs. $54\%$ \texttt{touching} relations). Each categorical attribute is mapped to numerical encodings, for example:
\textit{color}: red = 0, blue = 1, yellow = 2;
\textit{shape}: block = 0, wedge = 1, pyramid = 2.

\paragraph{Program Synthesis Data:}
For rule induction, we created a framework for generating rules inspired by the official Zendo card set by Looney Labs \footnote{\url{https://www.looneylabs.com/games/zendo}}. Each game pairs a DSL rule with 20 labeled examples (10 positive, 10 negative), generated via the Prolog pipeline.  The Rules take object vectors from \eqref{eq:vec} as input (omitting bounding boxes, unused for evaluation)

We categorize games by rule complexity:
\begin{description}
  \item[First Order:] Rules including predicates: \texttt{EXACTLY}, \texttt{AT\_LEAST}, \texttt{EXCLUSIVELY}, \texttt{ZERO}
  \item[Second Order:] Rules including predicates: \texttt{EVEN}, \texttt{ODD}, \texttt{MORE\_THAN}
  \item[Complex:] Rules describing relations (\texttt{POINTING}, \texttt{TOUCHING}, \texttt{ON\_TOP\_OF}) or combining multiple sub-rules using (\texttt{AND}, \texttt{OR}).
\end{description}
Additionally, we include a rule that includes the predicate \texttt{SAME\_AMOUNT} that acts as an out-of-distribution task for the Oracle Agent where we deliberately left out the predicate in its DSL.

We display all episodes in \autoref{app:tab:tasks} with the complexity level and the win rate per agent.

\begin{table*}[ht]
  \centering\small
  \setlength{\tabcolsep}{1.3pt}
  \caption{Per-rule win rates on all 28 tasks with \textbf{visual inputs}, one sub-column per backbone. Each cell shows wins/total runs over the seeds available for that task. Human participant results are wins/total players.}
  \label{app:tab:tasks}
  \resizebox{\textwidth}{!}{
  \begin{tabular}{p{3.3cm}lcccccccccccccc}
    \toprule
    \textbf{Rule} & \textbf{Complexity} & \multicolumn{4}{c}{\textbf{VLM Agent}} & \multicolumn{4}{c}{\textbf{Bayesian Agent}} & \multicolumn{4}{c}{\textbf{VLP Agent}} & \textbf{Oracle} & \textbf{Human} \\
    \cmidrule(lr){3-6}\cmidrule(lr){7-10}\cmidrule(lr){11-14}
     &  & {\scriptsize Mini} & {\scriptsize Terra} & {\scriptsize Gemma} & {\scriptsize Qwen} & {\scriptsize Mini} & {\scriptsize Terra} & {\scriptsize Gemma} & {\scriptsize Qwen} & {\scriptsize Mini} & {\scriptsize Terra} & {\scriptsize Gemma} & {\scriptsize Qwen} &  &  \\
    \midrule
    \texttt{\scriptsize At least 3 blocks or an odd number of flat pieces} & {\scriptsize Complex} & 0/5 & 0/5 & 0/5 & 0/5 & 0/5 & 0/5 & 0/5 & 0/5 & 0/5 & 0/5 & 0/5 & 0/5 & 5/5 & - \\
    \texttt{\scriptsize At least 3 blue pieces} & {\scriptsize First Order} & 5/5 & 5/5 & 0/5 & 0/5 & 4/5 & 5/5 & 4/5 & 2/5 & 3/5 & 4/5 & 0/5 & 0/5 & 5/5 & - \\
    \texttt{\scriptsize An even number of red pieces pointing at flat pieces} & {\scriptsize Complex} & 0/5 & 0/5 & 0/5 & 0/5 & 0/5 & 0/5 & 0/5 & 0/5 & 0/5 & 0/5 & 0/5 & 0/5 & 5/5 & - \\
    \texttt{\scriptsize At least 3 red pieces and exactly 1 pyramid} & {\scriptsize Complex} & 0/5 & 0/5 & 0/5 & 0/5 & 0/5 & 0/5 & 0/5 & 0/5 & 0/5 & 0/5 & 0/5 & 0/5 & 5/5 & 5/10 \\
    \texttt{\scriptsize More wedges than cheesecakes} & {\scriptsize Second Order} & 1/5 & 0/5 & 0/5 & 0/5 & 0/5 & 0/5 & 0/5 & 0/5 & 0/5 & 0/5 & 0/5 & 0/5 & 5/5 & - \\
    \texttt{\scriptsize At least 1 yellow piece on top of cheesecake} & {\scriptsize Complex} & 4/5 & 0/5 & 0/5 & 0/5 & 0/5 & 0/5 & 0/5 & 0/5 & 0/5 & 0/5 & 0/5 & 0/5 & 5/5 & 8/10 \\
    \texttt{\scriptsize Exactly 1 yellow flat piece} & {\scriptsize First Order} & 0/5 & 0/5 & 0/5 & 0/5 & 1/5 & 2/5 & 0/5 & 4/5 & 0/5 & 1/5 & 0/5 & 0/5 & 5/5 & - \\
    \texttt{\scriptsize An even number of blocks} & {\scriptsize Second Order} & 2/5 & 0/5 & 0/5 & 0/5 & 0/5 & 0/5 & 0/5 & 0/5 & 0/5 & 5/5 & 5/5 & 5/5 & 5/5 & - \\
    \texttt{\scriptsize An odd number of flat wedges} & {\scriptsize Second Order} & 0/5 & 0/5 & 0/5 & 0/5 & 0/5 & 0/5 & 0/5 & 0/5 & 0/5 & 0/5 & 0/5 & 0/5 & 5/5 & 5/10 \\
    \texttt{\scriptsize Either 2 or 1 pieces in total} & {\scriptsize Complex} & 5/5 & 5/5 & 2/5 & 1/5 & 0/5 & 4/5 & 0/5 & 0/5 & 1/5 & 4/5 & 0/5 & 0/5 & 5/5 & - \\
    \texttt{\scriptsize Odd number of yellow pieces touching wedges} & {\scriptsize Complex} & 0/5 & 0/5 & 0/5 & 0/5 & 0/5 & 0/5 & 0/5 & 0/5 & 0/5 & 0/5 & 0/5 & 0/5 & 5/5 & - \\
    \texttt{\scriptsize Even number of blue upright pieces} & {\scriptsize Second Order} & 0/5 & 0/5 & 0/5 & 0/5 & 0/5 & 0/5 & 0/5 & 0/5 & 0/5 & 4/5 & 0/5 & 0/5 & 5/5 & - \\
    \texttt{\scriptsize At least 3 yellow blocks} & {\scriptsize First Order} & 4/5 & 5/5 & 0/5 & 3/5 & 4/5 & 5/5 & 3/5 & 5/5 & 0/5 & 1/5 & 0/5 & 0/5 & 5/5 & 10/10 \\
    \texttt{\scriptsize Exactly 3 flat pieces} & {\scriptsize First Order} & 0/5 & 0/5 & 0/5 & 0/5 & 0/5 & 2/5 & 0/5 & 1/5 & 0/5 & 0/5 & 0/5 & 0/5 & 5/5 & - \\
    \texttt{\scriptsize Same number of blue pieces and red pieces} & {\scriptsize OOD} & 0/5 & 5/5 & 0/5 & 0/5 & 0/5 & 0/5 & 0/5 & 0/5 & 0/5 & 3/5 & 0/5 & 0/5 & 0/5 & 7/10 \\
    \texttt{\scriptsize All three shapes are present} & {\scriptsize First Order} & 4/5 & 0/5 & 2/5 & 0/5 & 0/5 & 0/5 & 1/5 & 1/5 & 0/5 & 0/5 & 0/5 & 0/5 & 5/5 & 9/10 \\
    \texttt{\scriptsize All three colors are present} & {\scriptsize First Order} & 5/5 & 5/5 & 4/5 & 4/5 & 2/5 & 4/5 & 5/5 & 1/5 & 0/5 & 0/5 & 0/5 & 0/5 & 5/5 & - \\
    \texttt{\scriptsize Only upright pieces} & {\scriptsize First Order} & 4/5 & 4/5 & 3/5 & 3/5 & 3/5 & 4/5 & 1/5 & 3/5 & 3/5 & 3/5 & 0/5 & 0/5 & 5/5 & - \\
    \texttt{\scriptsize No pyramids} & {\scriptsize First Order} & 5/5 & 3/5 & 0/5 & 1/5 & 1/5 & 3/5 & 4/5 & 1/5 & 3/5 & 3/5 & 0/5 & 1/5 & 5/5 & - \\
    \texttt{\scriptsize Odd number of total pieces} & {\scriptsize Second Order} & 3/5 & 5/5 & 0/5 & 0/5 & 0/5 & 0/5 & 0/5 & 0/5 & 5/5 & 5/5 & 5/5 & 5/5 & 5/5 & - \\
    \texttt{\scriptsize Even number of total pieces} & {\scriptsize Second Order} & 4/5 & 5/5 & 0/5 & 0/5 & 0/5 & 0/5 & 0/5 & 0/5 & 5/5 & 5/5 & 5/5 & 5/5 & 5/5 & - \\
    \texttt{\scriptsize Odd number of doorstops} & {\scriptsize Second Order} & 3/5 & 0/5 & 0/5 & 0/5 & 0/5 & 0/5 & 0/5 & 0/5 & 0/5 & 3/5 & 0/5 & 0/5 & 5/5 & - \\
    \texttt{\scriptsize More flat pieces than vertical pieces} & {\scriptsize Second Order} & 0/5 & 0/5 & 0/5 & 0/5 & 0/5 & 0/5 & 0/5 & 0/5 & 0/5 & 2/5 & 0/5 & 0/5 & 5/5 & - \\
    \texttt{\scriptsize At least 3 grounded yellow pieces} & {\scriptsize Complex} & 0/5 & 0/5 & 0/5 & 0/5 & 1/5 & 4/5 & 0/5 & 0/5 & 0/5 & 2/5 & 0/5 & 0/5 & 5/5 & - \\
    \texttt{\scriptsize Exactly 2 blue pieces} & {\scriptsize First Order} & 1/5 & 5/5 & 0/5 & 0/5 & 4/5 & 5/5 & 5/5 & 4/5 & 5/5 & 5/5 & 0/5 & 0/5 & 5/5 & - \\
    \texttt{\scriptsize Odd number of upside down yellow pieces} & {\scriptsize Second Order} & 0/5 & 0/5 & 0/5 & 0/5 & 0/5 & 0/5 & 0/5 & 0/5 & 0/5 & 0/5 & 0/5 & 0/5 & 5/5 & - \\
    \texttt{\scriptsize At least 2 blocks pointing at blue pieces} & {\scriptsize Complex} & 0/5 & 0/5 & 0/5 & 0/5 & 0/5 & 0/5 & 0/5 & 0/5 & 0/5 & 0/5 & 0/5 & 0/5 & 4/5 & - \\
    \texttt{\scriptsize Exactly 2 doorstops on top of wedges} & {\scriptsize Complex} & 0/5 & 0/5 & 0/5 & 0/5 & 0/5 & 0/5 & 0/5 & 0/5 & 0/5 & 0/5 & 0/5 & 0/5 & 0/5 & - \\
    \bottomrule
  \end{tabular}
  }
\end{table*}

\begin{table*}[ht]
  \centering\small
  \setlength{\tabcolsep}{1.3pt}
  \caption{Per-rule win rates on all 28 tasks with \textbf{symbolic inputs}, one sub-column per backbone. Each cell shows wins/total runs over the seeds available for that task.}
  \label{app:tab:symbolic_tasks}
  \resizebox{\textwidth}{!}{
    \begin{tabular}{p{3.3cm}lcccccccccccccc}
    \toprule
    \textbf{Rule} & \textbf{Complexity} & \multicolumn{4}{c}{\textbf{VLM Agent}} & \multicolumn{4}{c}{\textbf{Bayesian Agent}} & \multicolumn{4}{c}{\textbf{VLP Agent}} & \textbf{Oracle} & \textbf{Random} \\
    \cmidrule(lr){3-6}\cmidrule(lr){7-10}\cmidrule(lr){11-14}
     &  & {\scriptsize Mini} & {\scriptsize Terra} & {\scriptsize Gemma} & {\scriptsize Qwen} & {\scriptsize Mini} & {\scriptsize Terra} & {\scriptsize Gemma} & {\scriptsize Qwen} & {\scriptsize Mini} & {\scriptsize Terra} & {\scriptsize Gemma} & {\scriptsize Qwen} &  &  \\
    \midrule
    \texttt{\scriptsize At least 3 blocks or an odd number of flat pieces} & {\scriptsize Complex} & 0/5 & 0/5 & 0/5 & 0/5 & 0/5 & 0/5 & 0/5 & 0/5 & 0/5 & 0/5 & 0/5 & 0/5 & 5/5 & 5/5 \\
    \texttt{\scriptsize At least 3 blue pieces} & {\scriptsize First Order} & 5/5 & 5/5 & 4/5 & 0/5 & 5/5 & 5/5 & 3/5 & 3/5 & 5/5 & 3/5 & 0/5 & 2/5 & 5/5 & 5/5 \\
    \texttt{\scriptsize An even number of red pieces pointing at flat pieces} & {\scriptsize Complex} & 0/5 & 0/5 & 0/5 & 0/5 & 0/5 & 0/5 & 0/5 & 0/5 & 0/5 & 0/5 & 0/5 & 0/5 & 5/5 & 5/5 \\
    \texttt{\scriptsize At least 3 red pieces and exactly 1 pyramid} & {\scriptsize Complex} & 2/5 & 2/5 & 0/5 & 0/5 & 0/5 & 0/5 & 0/5 & 0/5 & 0/5 & 0/5 & 0/5 & 0/5 & 5/5 & 4/5 \\
    \texttt{\scriptsize More wedges than cheesecakes} & {\scriptsize Second Order} & 1/5 & 2/5 & 4/5 & 0/5 & 0/5 & 0/5 & 0/5 & 0/5 & 0/5 & 0/5 & 0/5 & 1/5 & 5/5 & 5/5 \\
    \texttt{\scriptsize At least 1 yellow piece on top of cheesecake} & {\scriptsize Complex} & 4/5 & 5/5 & 5/5 & 0/5 & 0/5 & 0/5 & 0/5 & 0/5 & 2/5 & 2/5 & 5/5 & 0/5 & 5/5 & 5/5 \\
    \texttt{\scriptsize Exactly 1 yellow flat piece} & {\scriptsize First Order} & 0/5 & 0/5 & 1/5 & 0/5 & 4/5 & 4/5 & 3/5 & 4/5 & 3/5 & 3/5 & 4/5 & 4/5 & 5/5 & 5/5 \\
    \texttt{\scriptsize An even number of blocks} & {\scriptsize Second Order} & 1/5 & 2/5 & 0/5 & 0/5 & 0/5 & 0/5 & 0/5 & 0/5 & 5/5 & 5/5 & 5/5 & 5/5 & 5/5 & 5/5 \\
    \texttt{\scriptsize An odd number of flat wedges} & {\scriptsize Second Order} & 0/5 & 0/5 & 0/5 & 0/5 & 0/5 & 0/5 & 0/5 & 0/5 & 2/5 & 2/5 & 5/5 & 5/5 & 5/5 & 5/5 \\
    \texttt{\scriptsize Either 2 or 1 pieces in total} & {\scriptsize Complex} & 5/5 & 5/5 & 5/5 & 0/5 & 0/5 & 1/5 & 0/5 & 0/5 & 4/5 & 5/5 & 0/5 & 0/5 & 5/5 & 5/5 \\
    \texttt{\scriptsize Odd number of yellow pieces touching wedges} & {\scriptsize Complex} & 0/5 & 0/5 & 0/5 & 0/5 & 0/5 & 0/5 & 0/5 & 0/5 & 0/5 & 0/5 & 5/5 & 0/5 & 5/5 & 5/5 \\
    \texttt{\scriptsize Even number of blue upright pieces} & {\scriptsize Second Order} & 2/5 & 4/5 & 0/5 & 0/5 & 0/5 & 0/5 & 0/5 & 0/5 & 4/5 & 5/5 & 1/5 & 5/5 & 5/5 & 5/5 \\
    \texttt{\scriptsize At least 3 yellow blocks} & {\scriptsize First Order} & 5/5 & 5/5 & 5/5 & 1/5 & 4/5 & 5/5 & 5/5 & 4/5 & 5/5 & 5/5 & 5/5 & 5/5 & 5/5 & 5/5 \\
    \texttt{\scriptsize Exactly 3 flat pieces} & {\scriptsize First Order} & 0/5 & 0/5 & 0/5 & 0/5 & 0/5 & 0/5 & 0/5 & 0/5 & 2/5 & 2/5 & 5/5 & 5/5 & 5/5 & 5/5 \\
    \texttt{\scriptsize Same number of blue pieces and red pieces} & {\scriptsize OOD} & 0/5 & 2/5 & 0/5 & 0/5 & 0/5 & 0/5 & 0/5 & 0/5 & 4/5 & 5/5 & 5/5 & 5/5 & 0/5 & 0/5 \\
    \texttt{\scriptsize All three shapes are present} & {\scriptsize First Order} & 5/5 & 5/5 & 5/5 & 0/5 & 2/5 & 0/5 & 0/5 & 0/5 & 0/5 & 0/5 & 0/5 & 0/5 & 5/5 & 5/5 \\
    \texttt{\scriptsize All three colors are present} & {\scriptsize First Order} & 5/5 & 5/5 & 5/5 & 0/5 & 5/5 & 1/5 & 4/5 & 0/5 & 0/5 & 0/5 & 0/5 & 0/5 & 5/5 & 5/5 \\
    \texttt{\scriptsize Only upright pieces} & {\scriptsize First Order} & 5/5 & 5/5 & 5/5 & 0/5 & 5/5 & 5/5 & 0/5 & 2/5 & 4/5 & 5/5 & 5/5 & 3/5 & 5/5 & 5/5 \\
    \texttt{\scriptsize No pyramids} & {\scriptsize First Order} & 5/5 & 5/5 & 5/5 & 2/5 & 1/5 & 3/5 & 0/5 & 0/5 & 5/5 & 5/5 & 5/5 & 5/5 & 5/5 & 5/5 \\
    \texttt{\scriptsize Odd number of total pieces} & {\scriptsize Second Order} & 5/5 & 5/5 & 0/5 & 0/5 & 0/5 & 0/5 & 0/5 & 0/5 & 5/5 & 5/5 & 5/5 & 5/5 & 5/5 & 5/5 \\
    \texttt{\scriptsize Even number of total pieces} & {\scriptsize Second Order} & 5/5 & 5/5 & 0/5 & 0/5 & 0/5 & 0/5 & 0/5 & 0/5 & 5/5 & 5/5 & 5/5 & 5/5 & 5/5 & 5/5 \\
    \texttt{\scriptsize Odd number of doorstops} & {\scriptsize Second Order} & 4/5 & 5/5 & 0/5 & 0/5 & 0/5 & 0/5 & 0/5 & 0/5 & 3/5 & 5/5 & 5/5 & 2/5 & 5/5 & 5/5 \\
    \texttt{\scriptsize More flat pieces than vertical pieces} & {\scriptsize Second Order} & 0/5 & 0/5 & 0/5 & 0/5 & 0/5 & 0/5 & 0/5 & 0/5 & 0/5 & 0/5 & 0/5 & 0/5 & 5/5 & 5/5 \\
    \texttt{\scriptsize At least 3 grounded yellow pieces} & {\scriptsize Complex} & 0/5 & 0/5 & 0/5 & 0/5 & 2/5 & 3/5 & 0/5 & 0/5 & 0/5 & 2/5 & 2/5 & 0/5 & 5/5 & 5/5 \\
    \texttt{\scriptsize Exactly 2 blue pieces} & {\scriptsize First Order} & 0/5 & 5/5 & 0/5 & 0/5 & 4/5 & 4/5 & 4/5 & 2/5 & 5/5 & 5/5 & 5/5 & 4/5 & 5/5 & 5/5 \\
    \texttt{\scriptsize Odd number of upside down yellow pieces} & {\scriptsize Second Order} & 1/5 & 4/5 & 3/5 & 0/5 & 0/5 & 1/5 & 0/5 & 0/5 & 5/5 & 5/5 & 5/5 & 5/5 & 5/5 & 5/5 \\
    \texttt{\scriptsize At least 2 blocks pointing at blue pieces} & {\scriptsize Complex} & 1/5 & 4/5 & 0/5 & 0/5 & 0/5 & 0/5 & 0/5 & 0/5 & 0/5 & 0/5 & 0/5 & 0/5 & 5/5 & 5/5 \\
    \texttt{\scriptsize Exactly 2 doorstops on top of wedges} & {\scriptsize Complex} & 0/5 & 2/5 & 0/5 & 0/5 & 0/5 & 0/5 & 0/5 & 0/5 & 1/5 & 2/5 & 2/5 & 3/5 & 5/5 & 5/5 \\
    \bottomrule
  \end{tabular}
    }
\end{table*}

\paragraph{Rule Guess Parsing.} \label{pg:rule_guessing}
When an agent's hypothesis cannot be parsed into a valid DSL program, the system falls back to querying an LLM for logical equivalence and counterexample generation instead of using the algorithmic comparison. This fallback is rare across all agents: the VLP Agent triggered it in 68 of 1,083 guesses (6.3\%), the VLM Agent in 34 of 665 (5.1\%) and the Bayesian Agent in 18 of 1,020 (1.8\%). The low fallback rates indicate that the evaluation pipeline relies on formal program comparison in the vast majority of cases. As the translation and evaluation rely on correctness of the backbone models answers used, we manually audited all resulting game files before the analysis. During manual inspection, we identified few evaluation errors, including an incorrect rule acceptance and cases where correct natural-language guesses were translated into incorrect DSL programs. We addressed these through a game rerun or by truncating the recorded interaction at the first correct guess.

\section{Technical Details}
\label{sec:Model}
The base of the Neuro-Symbolic Zendo player consists of two separate modules that are connected through a shared encoding of outputs and inputs. The first is a vision model that detects each piece in a Zendo structure. This model is trained on synthetic images generated specifically for the task. Its output is a list of vectors, each representing one piece in the structure. The second module is a program synthesizer based on DeepSynth \cite{fijalkow2022scaling}. Its inputs are input-output-pairs (I/O examples) in the form of lists of vectors and labels. The labels indicate whether a specific structure follows the hidden rule. Given several I/O examples, the synthesizer produces a program that classifies structures accordingly. This program can be interpreted directly as a logical rule.

\subsection{Vision Model}
The vision module detects the objects in a Zendo scene and predicts their attributes and spatial relations. Unlike bounding-box–based detectors, which struggle to represent relations such as touching or pointing, our model outputs a structured symbolic scene description suitable for program-based rule induction.

We use a ResNet-18 backbone followed by a small transformer encoder (4 layers) that produces a fixed set of T=7 object tokens. Each token is decoded by dedicated heads predicting object attributes (color, shape, orientation), presence, bounding boxes, and relational properties (touching and pointing). Relational heads output links between object indices, allowing the model to represent directed and undirected interactions between pieces.
The touching head outputs six values, each representing the direction in which a piece is touching another piece. The directions are ordered as: left, right, front, back, top, bottom. This is therefore used for the "TOUCHING" and "ON\_TOP\_OF" predicates in the program synthesis.

Training uses a permutation-invariant loss based on the Hungarian algorithm. Predicted and ground-truth objects are matched using bounding-box distance, and losses for attributes, relations, presence, and bounding boxes are computed on the aligned objects. The final objective is a weighted sum of these components.

This architecture produces a complete, symbolic scene encoding that feeds directly into the program-synthesis module.

\subsubsection{Perception Evaluation}
We evaluate the vision module on a synthetic test set of 2{,}779 images. After
Hungarian matching between predicted and ground-truth objects, the model achieves 
very high accuracy across all prediction heads, including color (98.9\%), shape 
(98.8\%), orientation (99.0\%), touching (98.4\%), pointing (96.9\%), and bounding boxes 
(99.3\%). The model also recovers the correct number of pieces in 99.8\% of scenes.
\autoref{tab:accuracy_loss_summary} summarizes these results.
Qualitative inspection confirms that both object attributes and relations are extracted
correctly.
\begin{table}[h!]
\centering
\renewcommand{\arraystretch}{1.15}
\caption{Accuracy and loss summary for each prediction head.}
\begin{tabular}{|l|c|c|c|}
\hline
\textbf{Head} & \textbf{Absolute Correct} & \textbf{Percentage Correct} & \textbf{Loss} \\
\hline
color        & 8935/9032 & 98.93\% & 0.0234 \\
shape        & 8919/9032 & 98.75\% & 0.0277 \\
orientation  & 8944/9032 & 99.03\% & 0.0258 \\
pointing     & 8752/9032 & 96.90\% & 0.0662 \\
bbox         & 8964/9032 & 99.25\% & 1.9853 \\
touching     & 8885/9032 & 98.37\% & 0.0375 \\
length       & 2773/2779 & 99.78\% & 0.0027 \\
\hline
\textbf{Total} & -- & -- & 0.0423 \\
\hline
\end{tabular}
\label{tab:accuracy_loss_summary}
\end{table}

Ablations confirm that transformer layers and dedicated relational heads are needed for 
generalization: removing transformer layers increases the aggregated validation loss to 1.76, compared to 0.0423 for the full model, indicating a substantial degradation in relational prediction. Using 
simpler relational heads yields a higher loss of 0.087.

\subsection{Program Synthesis} \label{app:dsl_details}
To infer Zendo rules from labeled examples, we use the DeepSynth framework \cite{fijalkow2022scaling}, which supports program induction over a DSL. The DSL is given in \autoref{app:grammar} which allows Zendo rules to be represented as executable logical programs . The DSL is directly aligned with the predicates used in data generation, ensuring consistency between the program synthesis component and the Prolog-based example generator.

DeepSynth converts this DSL into a probabilistic context-free grammar (PCFG) over rule programs. In principle, the PCFG probabilities can either be uniform or predicted by neural networks that condition on the input–output examples. In the uniform setting, all productions are assigned equal weight, and synthesis reduces to enumerating programs by increasing size under a flat prior. In the learned setting, neural predictors estimate a non-uniform distribution over productions, biasing search toward plausible hypotheses.

In the neuro-symbolic agent reported in the main paper, we use a uniform PCFG without neural predictors. This choice ensures a more even comparison to VLM-based agents, which do not have access to our training task distribution, and avoids leaking prior knowledge about Zendo rules into the search prior. The program synthesis module therefore operates as a purely symbolic enumerative search conditioned only on the observed examples.

Program induction is performed using heap search, which enumerates candidate programs in decreasing PCFG probability and checks their consistency against all labeled examples. The search procedure is loss-optimal with respect to the PCFG and can be instantiated with either uniform or learned probabilities.
\subsubsection{Grammar}
\paragraph{Types}
\label{app:grammar}

\begin{align*}
\mathsf{Piece} &::= \texttt{piece} \\
\mathsf{Structure} &::= \mathsf{List}(\mathsf{Piece}) \\
\mathsf{UnaryPred} &::= \mathsf{Piece} \to \mathsf{Bool} \\
\mathsf{InteractionPred} &::= \mathsf{Piece} \to (\mathsf{Structure} \to \mathsf{Bool}) \\
\mathsf{Rule} &::= \mathsf{Structure} \to \mathsf{Bool}
\end{align*}
\begin{align*}
\mathsf{UnaryPred} ::=~&
    \texttt{IS\_GROUNDED}
    \mid \texttt{IS\_UNGROUNDED}
    \mid \texttt{IS\_RED} \mid \texttt{IS\_BLUE} \mid \texttt{IS\_YELLOW} \\
&\mid \texttt{IS\_BLOCK} \mid \texttt{IS\_WEDGE} \mid \texttt{IS\_PYRAMID} \mid \texttt{IS\_UPRIGHT} \\
&\mid \texttt{IS\_UPSIDE\_DOWN} \mid \texttt{IS\_DOORSTOP} \mid \texttt{IS\_CHEESECAKE} \\
&\mid \texttt{IS\_VERTICAL} \mid \texttt{IS\_FLAT}
\end{align*}
\begin{align*}
\mathsf{InteractionPred} ::=~&
    \texttt{TOUCHING}~\mathsf{UnaryPred}~\mathsf{UnaryPred} \mid \texttt{POINTING}~\mathsf{UnaryPred}~\mathsf{UnaryPred} \\
&\mid \texttt{ON\_TOP\_OF}~\mathsf{UnaryPred}~\mathsf{UnaryPred}
\end{align*}
\paragraph{Syntax}
\label{app:dsl-syntax}

\begin{align*}
\mathsf{Rule} ::=~&
    \mathsf{UnaryRule}
    \mid \mathsf{BinaryRule}
    \mid \mathsf{InteractionRule} \\
&\mid \mathsf{SetRule}
   \mid \mathsf{ConnectiveRule}
\\[0.75em]
\mathsf{UnaryRule} ::=~&
    \texttt{AT\_LEAST\_1}~\mathsf{Int}~\mathsf{UnaryPred} \\
&\mid \texttt{EXACTLY\_1}~\mathsf{Int}~\mathsf{UnaryPred} \\
&\mid \texttt{ZERO\_1}~\mathsf{UnaryPred} \\
&\mid \texttt{EVEN\_1}~\mathsf{UnaryPred} \\
&\mid \texttt{ODD\_1}~\mathsf{UnaryPred} \\
&\mid \texttt{EXCLUSIVELY}~\mathsf{UnaryPred}
\\[0.75em]
\mathsf{BinaryRule} ::=~&
    \texttt{AT\_LEAST\_2}~\mathsf{Int}~\mathsf{UnaryPred}~\mathsf{UnaryPred} \\
&\mid \texttt{EXACTLY\_2}~\mathsf{Int}~\mathsf{UnaryPred}~\mathsf{UnaryPred} \\
&\mid \texttt{ZERO\_2}~\mathsf{UnaryPred}~\mathsf{UnaryPred} \\
&\mid \texttt{EVEN\_2}~\mathsf{UnaryPred}~\mathsf{UnaryPred} \\
&\mid \texttt{ODD\_2}~\mathsf{UnaryPred}~\mathsf{UnaryPred} \\
&\mid \texttt{SAME\_AMOUNT}~\mathsf{UnaryPred}~\mathsf{UnaryPred} \\
&\mid \texttt{MORE\_THAN}~\mathsf{UnaryPred}~\mathsf{UnaryPred}
\\[0.75em]
\mathsf{InteractionRule} ::=~&
    \texttt{AT\_LEAST\_INTERACTION}~\mathsf{Int}~\mathsf{InteractionPred} \\
&\mid \texttt{EXACTLY\_INTERACTION}~\mathsf{Int}~\mathsf{InteractionPred} \\
&\mid \texttt{EVEN\_INTERACTION}~\mathsf{InteractionPred} \\
&\mid \texttt{ODD\_INTERACTION}~\mathsf{InteractionPred}
\\[0.75em]
\mathsf{SetRule} ::=~&
    \texttt{EVEN} \mid \texttt{ODD} \\
&\mid \texttt{ALL\_THREE\_SHAPES} \\
&\mid \texttt{ALL\_THREE\_COLORS} \\
&\mid \texttt{EITHER\_OR}~\mathsf{Int}~\mathsf{Int}
\\[0.75em]
\mathsf{ConnectiveRule} ::=~&
    \texttt{AND}~\mathsf{Rule}~\mathsf{Rule} \mid \texttt{OR}~\mathsf{Rule}~\mathsf{Rule}
\\
\end{align*}

\section{Human Study Platform}\label{app:human_study}

\begin{figure}[h!]
  \centering
  \begin{subfigure}[t]{0.32\linewidth}
    \centering
    \includegraphics[width=\linewidth]{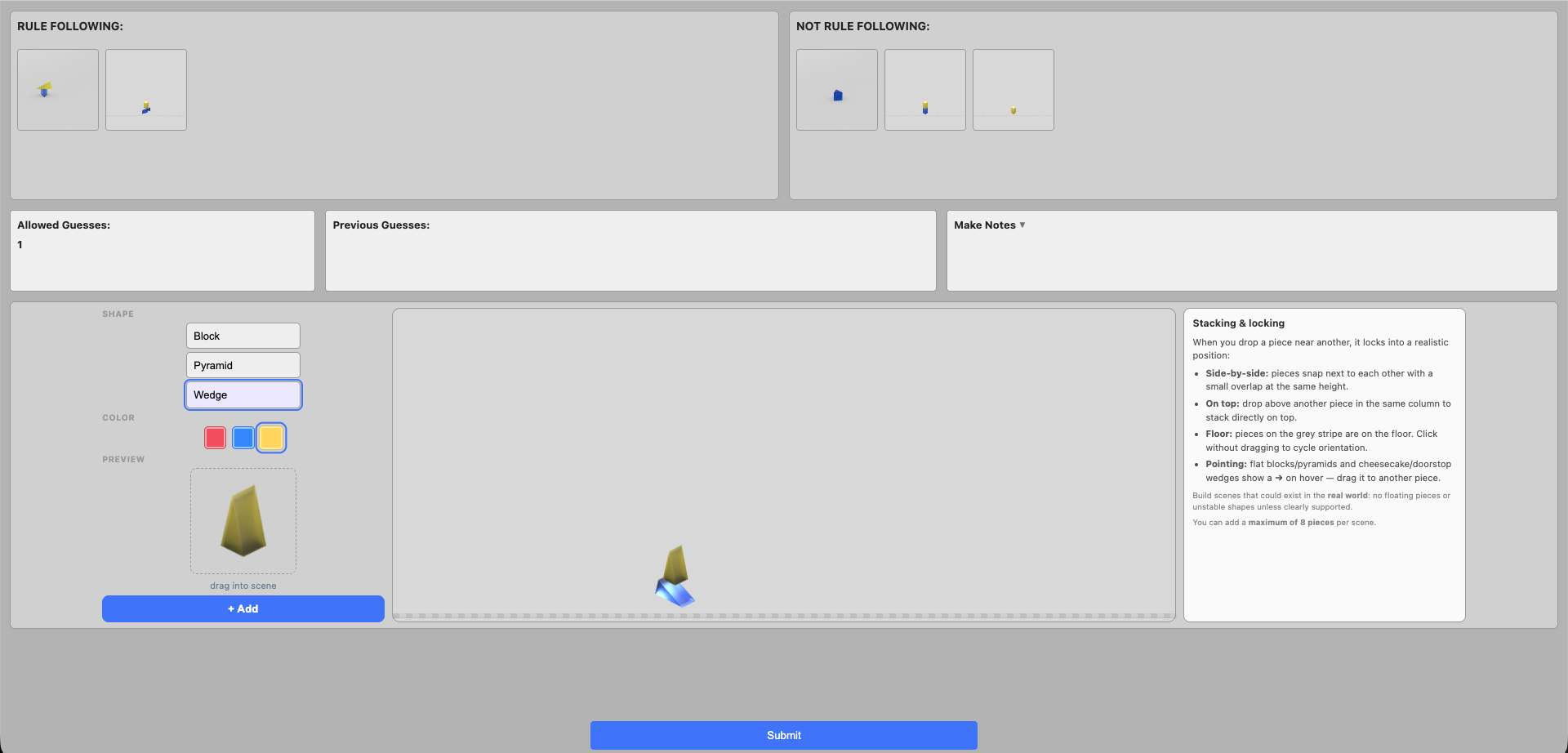}
    \caption{Build Scene Screen.}
    \label{fig:experiment}
  \end{subfigure}
  \hfill
  \begin{subfigure}[t]{0.32\linewidth}
    \centering
    \includegraphics[width=\linewidth]{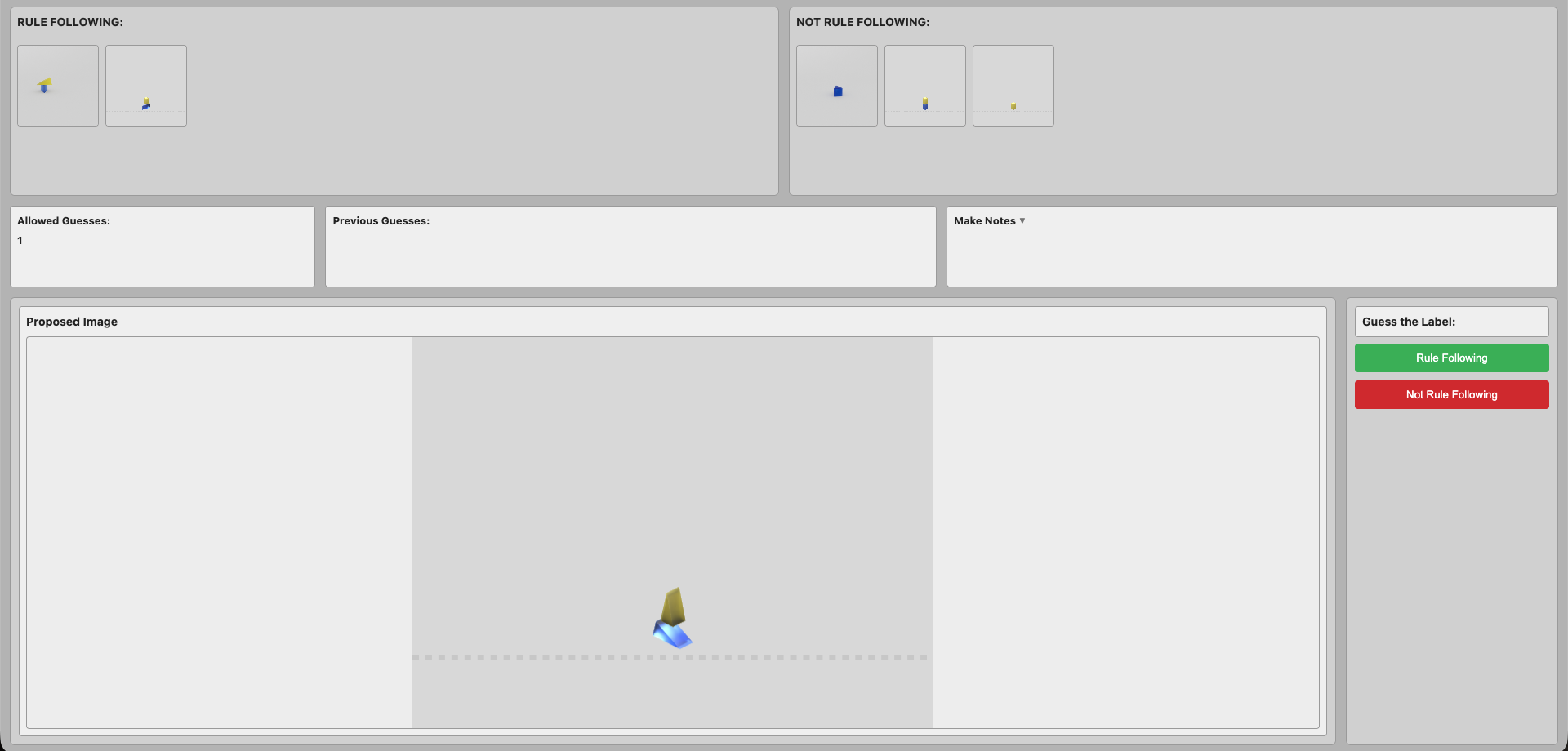}
    \caption{Guessing Label Screen.}
    \label{fig:label}
  \end{subfigure}
  \hfill
  \begin{subfigure}[t]{0.32\linewidth}
    \centering
    \includegraphics[width=\linewidth]{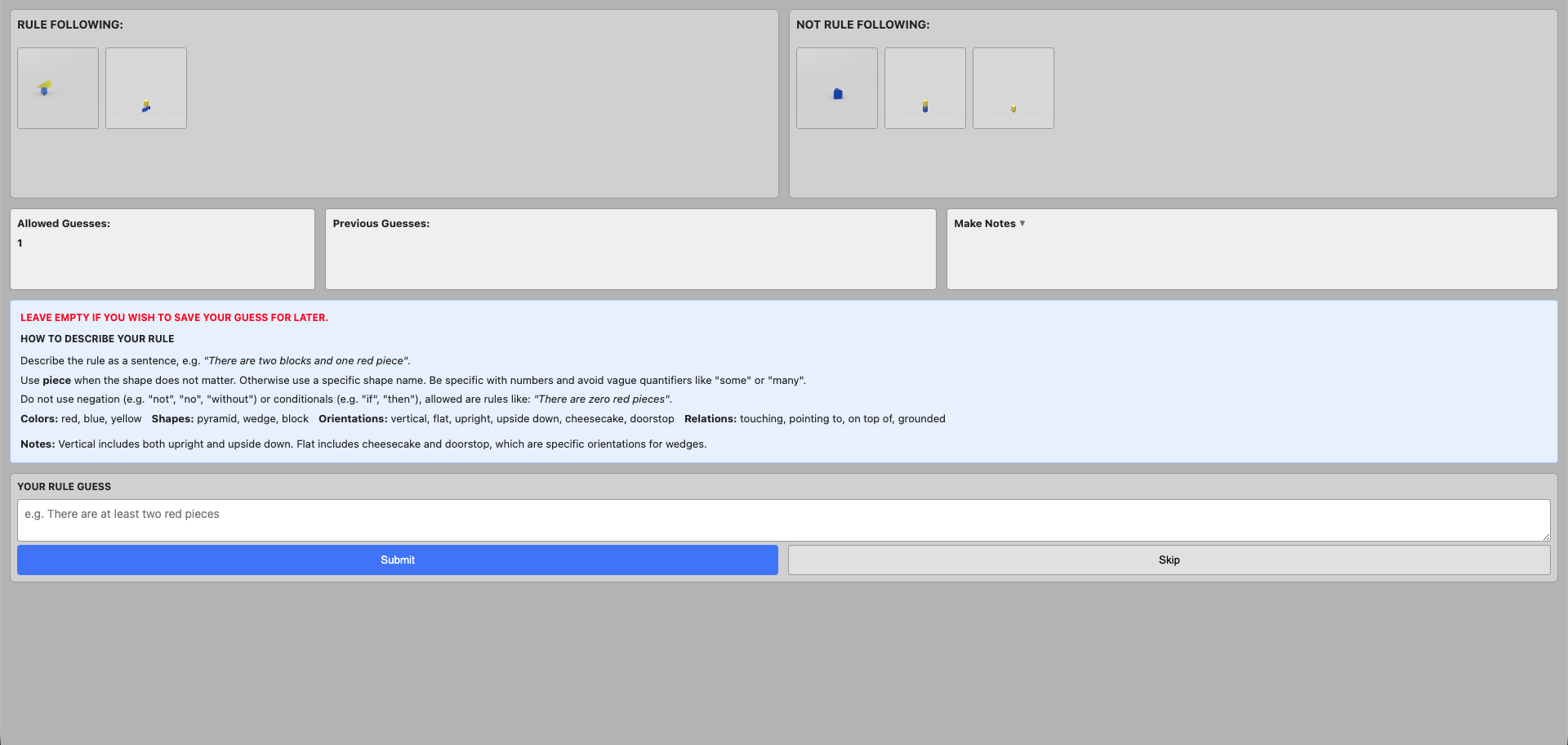}
    \caption{Guessing Rule Screen.}
    \label{fig:guess}
  \end{subfigure}
  \caption{Overview of study UI.}
  \label{fig:frontend}
\end{figure}

\subsection{Participants}
We collect data from 19 participants of which five played all six games, two played four games, two played three games, six played two games and four played just one game. Participants were recruited from within the institution. The participants play time varied between 10 to 40 minutes per game.

\subsection{System Architecture}

The human study is delivered through a purpose-built web application. The frontend is a React + TypeScript single-page application served over HTTPS. The backend is a Python FastAPI service that hosts the game logic and communicates with the frontend exclusively via WebSocket. The backend reuses the same GameMaster and DSL infrastructure as the agent experiments, ensuring that the rules, counterexample generation, and labeling logic are identical for human and AI players. The backend system is run with Docker and deployed on an internal university server and the frontend is deployed on a JATOS server.

\subsection{Study Procedure}

Participants access the study through a JATOS URL. The study proceeds in three phases:

\begin{enumerate}
    \item \textbf{Instructions and consent.} Participants read a study information sheet explaining the purpose, procedure, expected duration, data collected, storage, and their right to withdraw at any time. A required checkbox confirms informed consent before the study begins.
    \item \textbf{Tutorial.} Participants complete one full guided game in which on-screen annotations explain the interface elements and game rules. Tutorial data are excluded from analysis.
    \item \textbf{Study games.} Participants play a subset of the 28 \environment episodes following the same protocol as the AI Agents of the main paper. \autoref{fig:frontend} shows the three screens the user sees during the game. The first image (\autoref{fig:experiment}) is during the experimentation phase where the user can build the scene, the second image (\autoref{fig:label}) shows the user the scene and asks for a label and the third image (\autoref{fig:guess}) shows the rule guessing screen where the user can input their hypothesis.
\end{enumerate}

\subsection{Data Collection and Storage}

Two complementary log files are produced per episode and stored on university servers.

\paragraph{Action log.} A fine-grained event stream capturing every user interaction during scene construction and decision-making. Each entry contains an ISO 8601 timestamp and an action type together with action-specific details. Logged events include:

\begin{itemize}
    \item \textit{Scene building:} \texttt{piece\_added}, \texttt{piece\_orientation\_cycled}, \texttt{piece\_moved}, \texttt{piece\_touching\_set/removed}, \texttt{piece\_pointing\_set/removed}, \texttt{piece\_stacked\_set/removed}.
    \item \textit{Guessing:} \texttt{guess\_label} (YES/NO prediction), \texttt{quiz\_result} (correctness and stone count), \texttt{rule\_typed} (full text of hypothesis), \texttt{rule\_submitted}, \texttt{rule\_skipped}.
    \item \textit{Navigation:} \texttt{screen\_change}, \texttt{received\_labeled\_example}, \texttt{game\_started}, \texttt{game\_over}, \texttt{session\_complete}.
\end{itemize}

\paragraph{Game state file.} A structured JSON summary saved at episode completion containing: the ground-truth DSL rule, rule complexity, number of turns and examples, all rule hypotheses submitted, label-guess correctness per turn, per-turn durations in seconds, the game-over reason, and file paths to the rendered scene images.

\paragraph{Scene images.} For every scene shown to or proposed by a participant, a $640\times480$ PNG rendered by Blender is saved alongside the game state file. Images are named by source (\texttt{gm} for Game-Master-provided, \texttt{player} for participant-proposed) and ground-truth label.

\paragraph{JATOS submission.} Upon episode completion the frontend submits a result payload to the JATOS server via \texttt{jatos.submitResultData}. The payload contains the session metadata (anonymous participant ID, browser user-agent, screen resolution) and the complete action log for each completed episode.

\subsection{Anonymization and Privacy}

No personal data are collected at any point. Each participant is assigned an anonymous session ID of the form \texttt{s\_XXXXXXXX} (8-character hex string) generated client-side and stored in browser \texttt{localStorage}. All stored files and JATOS result entries are keyed by this ID. If JATOS is available, the JATOS worker ID is also recorded to enable session linkage within the JATOS platform, but this ID carries no personally identifiable information. No IP addresses or identifying metadata are logged by the application. The privacy notice presented to participants states explicitly that the data cannot be traced back to any individual.

\subsection{Participant Compensation and Voluntariness}

Participation is entirely voluntary and participants may withdraw at any time by closing the browser window without any negative consequences. No compensation is provided. Data from sessions that do not reach the study games phase are excluded from analysis.

\subsection{Instructions}
The following instructions are given to the participants before starting the games:
\small{\input{user_instructions}}

\section{Human Study Results}\label{app:results_human}
The results of the human study are reported in \autoref{tab:human_task_results}, alongside the per-task performance of the evaluated agents. The tasks that were most difficult for human participants, Tasks 3 and 8, were solved by only 5/10 participants and were also particularly challenging for the evaluated agents, with none of the non-oracle agents successfully completing either task. In contrast, Task 12 was solved by all human participants and was also solved by several agent configurations. Overall, these results suggest that tasks involving more complex rules are challenging for both humans and the evaluated methods. Nevertheless, even on the most difficult tasks, half of the human participants were able to find a solution within a relatively small number of turns. Notably, on Task 5, human participants who solved the task did so in fewer turns on average than the \textit{Oracle Agent} (6.2 versus 8.2 turns).
\begin{table*}[ht]
  \centering
  \small
  \setlength{\tabcolsep}{3pt}
  \caption{Per-task results on the 6 human-study tasks across backbones. Solved shows wins/total (seeds for agents, participants for humans). Avg Trn shows mean turns $\pm$ sample SD across winning runs only. -- = no winning runs; $n=1$ = one winner (SD undefined); zero SD is omitted. Oracle is backbone-independent.}
  \label{tab:human_task_results}
  \resizebox{\textwidth}{!}{
    \begin{tabular}{lcccccccccccc}
    \toprule
    \textbf{Player} & \multicolumn{2}{c}{\textbf{Task 3}} & \multicolumn{2}{c}{\textbf{Task 5}} & \multicolumn{2}{c}{\textbf{Task 8}} & \multicolumn{2}{c}{\textbf{Task 12}} & \multicolumn{2}{c}{\textbf{Task 14}} & \multicolumn{2}{c}{\textbf{Task 15}} \\
    \cmidrule(lr){2-3}\cmidrule(lr){4-5}\cmidrule(lr){6-7}\cmidrule(lr){8-9}\cmidrule(lr){10-11}\cmidrule(lr){12-13}
     & \textbf{Solved} & \textbf{Avg Trn} & \textbf{Solved} & \textbf{Avg Trn} & \textbf{Solved} & \textbf{Avg Trn} & \textbf{Solved} & \textbf{Avg Trn} & \textbf{Solved} & \textbf{Avg Trn} & \textbf{Solved} & \textbf{Avg Trn} \\
    \midrule
    Oracle Agent & 5/5 & 9.4 {\small $\pm$ 1.3} & 5/5 & 8.2 {\small $\pm$ 0.8} & 5/5 & 4.6 {\small $\pm$ 0.9} & 5/5 & 6.4 {\small $\pm$ 1.1} & 0/5 & -- & 5/5 & 2.2 {\small $\pm$ 0.4} \\
    \midrule
    VLM Agent (Mini) & 0/5 & -- & 4/5 & 12.0 {\small $\pm$ 3.2} & 0/5 & -- & 4/5 & 8.2 {\small $\pm$ 2.6} & 0/5 & -- & 4/5 & 6.2 {\small $\pm$ 1.5} \\
    Bayesian Agent (Mini) & 0/5 & -- & 0/5 & -- & 0/5 & -- & 4/5 & 6.2 {\small $\pm$ 1.0} & 0/5 & -- & 0/5 & -- \\
    VLP Agent (Mini) & 0/5 & -- & 0/5 & -- & 0/5 & -- & 0/5 & -- & 0/5 & -- & 0/5 & -- \\
    \midrule
    VLM Agent (Terra) & 0/5 & -- & 0/5 & -- & 0/5 & -- & 5/5 & 6.2 {\small $\pm$ 1.1} & 5/5 & 9.2 {\small $\pm$ 2.8} & 0/5 & -- \\
    Bayesian Agent (Terra) & 0/5 & -- & 0/5 & -- & 0/5 & -- & 5/5 & 8.6 {\small $\pm$ 2.6} & 0/5 & -- & 0/5 & -- \\
    VLP Agent (Terra) & 0/5 & -- & 0/5 & -- & 0/5 & -- & 1/5 & 17.0 {\small ($n=1$)} & 3/5 & 7.7 {\small $\pm$ 4.6} & 0/5 & -- \\
    \midrule
    VLM Agent (Gemma) & 0/5 & -- & 0/5 & -- & 0/5 & -- & 0/5 & -- & 0/5 & -- & 2/5 & 12.5 {\small $\pm$ 3.5} \\
    Bayesian Agent (Gemma) & 0/5 & -- & 0/5 & -- & 0/5 & -- & 3/5 & 11.3 {\small $\pm$ 2.1} & 0/5 & -- & 1/5 & 7.0 {\small ($n=1$)} \\
    VLP Agent (Gemma) & 0/5 & -- & 0/5 & -- & 0/5 & -- & 0/5 & -- & 0/5 & -- & 0/5 & -- \\
    \midrule
    VLM Agent (Qwen) & 0/5 & -- & 0/5 & -- & 0/5 & -- & 3/5 & 9.3 {\small $\pm$ 3.2} & 0/5 & -- & 0/5 & -- \\
    Bayesian Agent (Qwen) & 0/5 & -- & 0/5 & -- & 0/5 & -- & 5/5 & 9.2 {\small $\pm$ 1.1} & 0/5 & -- & 1/5 & 11.0 {\small ($n=1$)} \\
    VLP Agent (Qwen) & 0/5 & -- & 0/5 & -- & 0/5 & -- & 0/5 & -- & 0/5 & -- & 0/5 & -- \\
    \midrule
    Human & 5/10 & 12.2 {\small $\pm$ 5.0} & 8/10 & 6.2 {\small $\pm$ 2.5} & 5/10 & 9.6 {\small $\pm$ 3.2} & 10/10 & 7.6 {\small $\pm$ 2.8} & 7/10 & 8.3 {\small $\pm$ 3.7} & 9/10 & 6.8 {\small $\pm$ 2.7} \\
    \bottomrule
  \end{tabular}
  }
\end{table*}

\section{Agent Prompts}
\label{appendix:prompts}
During the hypothesis evaluation of the agents' guesses, we use the following prompt to convert the natural language hypotheses into processable DSL strings:
\begin{lstlisting}
    Convert the following hypothesis, from a different DSL about Zendo structures into a DSL representation using the provided syntax.
    Output only the DSL expression without any additional text. If the rule cannot be represented with the DSL, return something that is similar.
    
    Hypothesis: {hs}
    
    DSL Syntax:
    Use S-expressions: (OP arg1 arg2 ...) with prefix notation.
    A complete output must be a single DSL expression.
    
    --------------------------------------------------
    Unary predicates (no arguments)
    --------------------------------------------------
    Colors:
    - IS_RED
    - IS_BLUE
    - IS_YELLOW
    
    Shapes:
    - IS_BLOCK
    - IS_WEDGE
    - IS_PYRAMID
    
    Grounding:
    - IS_GROUNDED
    - IS_UNGROUNDED
    
    Orientation:
    - IS_UPRIGHT
    - IS_UPSIDE_DOWN
    - IS_VERTICAL
    - IS_FLAT
    - IS_DOORSTOP
    - IS_CHEESECAKE
    
    --------------------------------------------------
    Interaction predicates
    (take 2 unary predicates; return an interaction predicate)
    --------------------------------------------------
    - (TOUCHING P1 P2)
    - (ON_TOP_OF P1 P2)
    - (POINTING P1 P2)
    
    --------------------------------------------------
    Logical rule combinators
    --------------------------------------------------
    - (AND R1 R2)
    - (OR R1 R2)
    
    --------------------------------------------------
    Global rules (no arguments)
    --------------------------------------------------
    - (ALL_THREE_COLORS)
    - (ALL_THREE_SHAPES)
    - (EVEN)          ; total number of pieces is even
    - (ODD)           ; total number of pieces is odd
    
    --------------------------------------------------
    Count rules over unary predicates
    (INT n ∈ (1,2,3,4,5,6,7))
    --------------------------------------------------
    - (AT_LEAST_1 n P)
    - (EXACTLY_1 n P)
    - (ZERO_1 P)
    - (EVEN_1 P)      ; nonzero even count
    - (ODD_1 P)       ; odd count
    - (EXCLUSIVELY P) ; all pieces satisfy P
    - (ALL_1 P)       ; all pieces satisfy P
    - (MAJORITY_1 P)  ; at least half of pieces satisfy P
    
    --------------------------------------------------
    Count rules over conjunction on the same piece
    (INT n ∈ (1,2,3,4,5,6,7))
    --------------------------------------------------
    - (AT_LEAST_2 n P1 P2)
    - (EXACTLY_2 n P1 P2)
    - (ZERO_2 P1 P2)
    - (EVEN_2 P1 P2)
    - (ODD_2 P1 P2)
    - (ALL_2 P1 P2)        ; all pieces satisfy (P1 AND P2)
    - (MAJORITY_2 P1 P2)   ; at least half satisfy (P1 AND P2)
    
    --------------------------------------------------
    Count rules over interactions
    (INT n ∈ (1,2,3,4,5,6,7))
    --------------------------------------------------
    - (AT_LEAST_INTERACTION n (REL P1 P2))
    - (EXACTLY_INTERACTION n (REL P1 P2))
    - (EVEN_INTERACTION (REL P1 P2))
    - (ODD_INTERACTION (REL P1 P2))
    - (MAJORITY_INTERACTION (REL P1 P2))
    
    --------------------------------------------------
    Other numeric / comparison rules
    --------------------------------------------------
    - (LENGTH n)                 ; total number of pieces is exactly n
    - (EITHER_OR n1 n2)          ; total number of pieces is n1 or n2
    - (MORE_THAN P1 P2)          ; count(P1) > count(P2)
    - (MORE_OR_EQUAL_THAN P1 P2) ; count(P1) ≥ count(P2)
    - (SAME_AMOUNT P1 P2)        ; count(P1) == count(P2) and > 0
    
    --------------------------------------------------
    Examples / templates
    --------------------------------------------------
    Single attribute:
    - “at least 3 wedges”:
      (AT_LEAST_1 3 IS_WEDGE)
    
    Two attributes on same piece:
    - “exactly 2 red blocks”:
      (EXACTLY_2 2 IS_RED IS_BLOCK)
    
    Interaction count:
    - “at least one blue on top of a yellow”:
      (AT_LEAST_INTERACTION 1 (ON_TOP_OF IS_BLUE IS_YELLOW))
    
    Majority:
    - “most pieces are red”:
      (MAJORITY_1 IS_RED)
    
    Comparisons:
    - “more red pieces than blue pieces”:
      (MORE_THAN IS_RED IS_BLUE)
    
    Combine clauses:
    - (AND R1 R2)
    - (OR R1 R2)
    
    --------------------------------------------------
    
    Return only the DSL expression below inside a single code block, like this:
    ```python
    (...)
    ```
    
\end{lstlisting}
If this fails we use the following prompt to ask for equivalence:
\begin{lstlisting}
    We are playing the game Zendo with the following attributes for the pieces:
    - color: red, blue, yellow;
    - shape: block, wedge, pyramid;
    - orientation: upright, upside_down, flat, cheesecake, doorstop;
    - relation: grounded (whether the piece is touching the ground), touching(ID) (whether the piece is touching another piece with ID), pointing(ID) (whether the piece is pointing to another piece with ID), on_top_of(ID) (whether the piece is on top of another piece with ID).
    Wedges are never flat but instead can be doorstop or cheesecake, while the two other shapes can be flat but not cheesecake or doorstop.
    Interactions can be: grounded, touching(ID), pointing(ID) and on_top_of(ID), where ID is the first field of another piece, e.g. "pointing(2)" means this piece is pointing to the piece with ID 2.
    Is the first DSL rule semantically equivalent to the second DSL rule?
    
    Rule 1:
    {rule1}
    
    Rule 2:
    {rule2}
    
    ----------------------------------------
    Output format
    ----------------------------------------
    
    You MUST return exactly one of the following:
    
    Case 1 — Equivalent:
    ```json
    {{"equivalent": true}}
    ```
    Case 2 — Not equivalent:
    ```json
    {{
      "equivalent": false,
      "counterexample": "<structure description>"
    }}
    ```
    ----------------------------------------
    Counterexample requirements
    ----------------------------------------
    
    The counterexample evaluate differently on the two rules.
    
    Describe the structure using ONLY this format:
    
    piece 1: color = ..., shape = ..., orientation = ..., relation = ...
    
    piece 2: color = ..., shape = ..., orientation = ..., relation = ...
    ...
    
    Do NOT include explanations
    
    Do NOT include reasoning
    
    Do NOT include any text outside the JSON
    
    Return ONLY the JSON block.
\end{lstlisting}
Using the output from the equivalence check, we ask the LLM for a counter example using the following prompt:
\begin{lstlisting}
    A structure has one or more pieces/items. Each piece should contain the following attributes: 
    - color: red, blue, yellow;
    - shape: block, wedge, pyramid;
    - orientation: upright, upside_down, flat, cheesecake, doorstop;
    - relation: grounded (whether the piece is touching the ground), touching(ID) (whether the piece is touching another piece with ID), pointing(ID) (whether the piece is pointing to another piece with ID), on_top_of(ID) (whether the piece is on top of another piece with ID).
    Wedges are never flat but instead can be doorstop or cheesecake, while the two other shapes can be flat but not cheesecake or doorstop.
    Interactions can be: grounded, touching(ID), pointing(ID) and on_top_of(ID), where ID is the first field of another piece, e.g. "pointing(2)" means this piece is pointing to the piece with ID 2.
    
    Your task is to convert the following structure description into the wanted format.
    Description: {description}
    
    ONLY return a new example within a python block, in this exact format.
    Here are examples of valid formats:
    ["item(0, red, block, upright, grounded)", "item(1, blue, wedge, doorstop, touching(0))"]
    ["item(0, yellow, pyramid, upside_down, grounded)", "item(1, blue, wedge, doorstop, on_top_of(0))", "item(2, red, block, upright, pointing(1))"]
    Return your answer in this exact format including the python block:
    ```python
    ["item(ID, color, shape, orientation, interaction)", ...]
    ```
\end{lstlisting}

\subsection{VLM Agent}
The VLM Agent uses the following prompt to propose examples:
\begin{lstlisting}
    You are a Zendo player. Your goal is to **gain new information** about the hidden rule by proposing a \
    **novel** structure that is **maximally informative** (highly likely to change or confirm current beliefs).
    
    You are given *visual* positive and negative examples. Study the images, but output your proposal as **text**.
    The pieces can have
    - colors: red, blue, yellow;
    - shapes: block, wedge, pyramid;
    - orientations: upright, upside_down, flat, cheesecake, doorstop.
    Wedges are never flat but instead can be doorstop or cheesecake, while the two other shapes can be flat but not cheesecake or doorstop.
    Interactions can be: grounded, touching(ID), pointing(ID) and on_top_of(ID), where ID is the first field of another piece, e.g. "pointing(2)" means this piece is pointing to the piece with ID 2.
    
    DO NOT include explanations, reasoning, comments, or text outside the python block, in this exact format, where label is 1 for valid and 0 for invalid:
    Here are examples of valid answers:
    ```python
    [["item(0, red, block, upright, grounded)", "item(1, blue, wedge, doorstop, touching(0))"], 1]
    ```
    ```python
    [["item(0, yellow, pyramid, upside_down, grounded)", "item(1, blue, wedge, doorstop, on_top_of(0))", "item(2, red, block, upright, pointing(1))"], 0]
    ```
    Return your answer in this exact format including the python block:
    ```python
    [["item(ID, color, shape, orientation, interaction)", ...], label]
\end{lstlisting}
To guess rules, the Agent uses the following prompt:
\begin{lstlisting}
    You are a Zendo player. Your job is to find a new logical classification rule for given examples with labels. You are given a few positive and negative examples. Each image consists of pieces in different configurations.
    **Available values:**
    - Colors: red, blue, yellow
    - Shapes: block, wedge, pyramid
    - Orientations: upright, upside_down, flat, cheesecake, doorstop, vertical
    - Interactions: grounded, touching, pointing, on_top_of, groundedness
    
    **Goal:**
    Find **one** rule that is **True for all positive examples** and **False for all negative examples**.
    
    **Critical rules for selecting your answer:**
    1. You may combine short rules using "and" or "or" if needed, but **if a single predicate works, use it.**
    2. The rule must be as short and simple as possible but still accurate.
    3. Return **only** the rule - no explanation, no formatting, no extra text.
    4. Do not use any conditionals ("if", "when", "only if", etc.) or any text outside the rule itself.
    
    ### Output Format:
    Return **only** a single rule in natural language. Do **not** include explanations or extra text.
\end{lstlisting}

\subsection{VLP Agent}
The following prompts were used in the beginning of the game to discover predicates over the images:\\
Object Discovery:
\begin{lstlisting}
    You are analyzing images to identify the base shapes of objects that appear. Focus on the fundamental shape or kind of each object, not its color, size, or orientation - those are properties, not object types.
    
    IMPORTANT:
    - Return only atomic object type names (e.g., "block", "wedge", "pyramid", "sphere", "cube", "doorstop")
    - Do NOT include color, size, or orientation in the object name (e.g., "red block" is wrong, use "block")
    - Do NOT combine multiple attributes into one name (e.g., "yellow cheesecake" is wrong, use "cheesecake")
    - If the same shape appears in different colors or sizes, it is still the same object type
    - Return exactly {n} objects in a Python list
    
    Answer format:
    ```python
    objects = [...]
    ```
    No comments or explanations. If no objects found, return [].
\end{lstlisting}
Property Discovery:
\begin{lstlisting}
    # Property Discovery Task
    
    You are analyzing a set of images to identify important properties that describe objects in the image set.
    
    ## Objective
    
    Discover **atomic visual properties** that characterize objects in the images. Each property should be a single attribute, not a combination.
    
    ## Instructions
    
    1. **Examine all images carefully** - Look for properties that apply to the relevant objects across the image set
    2. **Identify important properties** - Focus on significant, clearly observable properties that meaningfully describe objects
    3. **Consider property variation** - Properties that vary across images or objects may be particularly noteworthy
    4. **Prioritize meaningful properties** - Choose properties that help distinguish or characterize objects (e.g., color, size, position, orientation, state)
    5. **Return exactly {n} properties** - If fewer notable properties exist, return as many as available
    6. **Use descriptive names** - Name properties clearly and specifically (e.g., "red" rather than "colored", "horizontal" rather than "oriented")
    
    ## Relevant Objects
    The objects to consider are: {objects}
    
    ## Property Categories
    - **Visual attributes**: colors (red, green, blue, yellow etc.)
    - **Geometric attributes**: orientations (upright, flat, upside_down etc.)
    - **Groundedness**: grounded, ungrounded
    
    ## Output Requirements
    - Return a Python list assigned to variable `properties`
    - Include only the Python code, no explanations or comments
    - If no notable properties are found, return an empty list `[]`
    - Use single-word or simple property names (e.g., "blue", "upright")
    - Do NOT include object types as properties (e.g., "wedge")
    
    ## Example Format
    ```python
    properties = ["red", "blue", "yellow", "upright", "flat", "doorstop", "grounded", "small"]
    ```
\end{lstlisting}
Action Discovery:
\begin{lstlisting}
    You are analyzing a set of images to identify important relations between objects.
    
    Discover **notable relations** that characterize the placement of two objects in the images. Focus on relations that meaningfully distinguish images.
    
    1. **Examine all images carefully** - Look for relations that apply to the relevant objects across the image set
    2. **Identify important relations** - Focus on significant, clearly observable actions that meaningfully describe what objects are doing
    3. **Consider relation variation** - Relations that vary across images or objects may be particularly noteworthy (contrasting actions)
    4. **Return exactly {n} relations** - If fewer notable relations exist, return as many as available
    5. **Use descriptive names** - Name relations clearly and specifically (e.g., "running" rather than "moving", "sitting" rather than "positioned", "touching" rather than "close")
    
    ## Relevant Objects
    
    The objects to consider are: {objects}
    
    ## Output Requirements
    
    - Return a Python list assigned to variable `actions`
    - Include only the Python code, no explanations or comments
    - If no notable actions are found, return an empty list `[]`
    - Use clear, specific action names
    
    ## Example Format
    ```python
    actions = ["touching", "pointing_to"]
    ```
\end{lstlisting}

The following prompts are used within the DSL programs evaluation to detect pieces with specific variables in the images:\\
Object detection:
\begin{lstlisting}
## Task
        Identify objects and their properties from the image using only the provided lists.

        **Objects:** {objects}  
        **Properties:** {properties}

        ## Rules
        1. Only use objects/properties from the provided lists
        2. Return empty list if no valid objects found
        3. No explanations or additional text

        ## Output Format
        ```python
        objects = [
            ['object_name', 'property1', 'property2', ...],
            ['object_name', 'property1'],
            ...
        ]
        ```

        **If no valid objects:** `objects = [[]]`

        ## Examples

        **Example 1**
        - Objects: ["car", "person", "tree"]
        - Properties: ["red", "tall", "small", "standing"]
        - Image: Red car under tall tree with small standing person

        ```python
        objects = [
            ['car', 'red'],
            ['tree', 'tall'], 
            ['person', 'standing', 'small']
        ]
        ```

        **Example 2**
        - Objects: ["dog", "ball", "book", "chair"] 
        - Properties: ["blue", "sitting", "round"]
        - Image: Dog sitting by round ball and blue chair

        ```python
        objects = [
            ['dog', 'sitting'],
            ['ball', 'round'],
            ['chair', 'blue']
        ]
        ```

        **Example 3**
        - Objects: ["bicycle", "lamp", "table", "cup"]
        - Properties: ["green", "broken", "wooden", "white"]
        - Image: Table with laptop and cup

        ```python
        objects = [[]]
        ```
        *Note: Even though 'table' and 'cup' are in the objects list and visible in the image, neither has properties from the provided list, so no valid object-property combinations exist*

        **Analyze the image now:**
\end{lstlisting}
Property detection:
\begin{lstlisting}
## Task
        Identify and extract properties for each object in the image using only the provided lists.

        **Properties:** {properties}

        ## Rules
        1. Only use properties from the provided lists
        2. Return empty list if no valid objects found
        3. No explanations or additional text

        ## Output Format
        ```python
        properties = [
            ['property1', 'property2', ...],
            ['property1'],
            ...
        ]
        ```

        **If no valid objects:** `objects = [[]]`

        ## Examples

        **Example 1**
        - Objects: ["car", "person", "tree"]
        - Properties: ["red", "tall", "small", "standing"]
        - Image: Red car under tall tree with small standing person

        ```python
        properties = [
            ['red'],
            ['tall'],
            ['standing', 'small']
        ]
        ```

        **Example 2**
        - Objects: ["dog", "ball", "book", "chair"] 
        - Properties: ["blue", "sitting", "round"]
        - Image: Dog sitting by round ball and blue chair

        ```python
        properties = [
            ['sitting'],
            ['round'],
            ['blue']
        ]
        ```

        **Example 3**
        - Properties: ["green", "broken", "wooden", "white"]
        - Image: Table with laptop and cup

        ```python
        properties = [[]]
        ```
        *Note: Even though 'table' and 'cup' are in the objects list and visible in the image, neither has properties from the provided list, so no valid object-property combinations exist*

        **Analyze the image now:**
\end{lstlisting}
Action detection:
\begin{lstlisting}
## Task
        Identify interaction triples in the image using only the provided lists.                                          
        Each triple has the format [subject, action, object] where:                                                       
        - subject: the entity performing the action (from Objects or Properties)                                          
        - action: what is happening between them (from Actions)                                                           
        - object: the entity the action is directed at (from Objects or Properties) 

        **Objects:** {objects}
        **Properties:** {properties}                                                                                      
        **Actions:** {actions}  

        ## Rules
        1. Only use values from the provided lists for each position 
        2. Each triple must have exactly 3 elements: [subject, action, object]  
        3. If the same interaction occurs multiple times in the image, include one entry per occurrence 
        4. Return empty list if no valid triples found  
        5. No explanations or additional text 

        ## Output Format
        ```python
        actions = [
            ['subject1', 'action1', 'object1'], 
            ['subject2', 'action2', 'object2'],
            ...
        ]
        ```

        **If no valid interactions:** `actions = [[]]`

        ## Examples

        **Example 1**
        - Objects: ["block", "pyramid", "wedge"] 
        - Properties: ["blue", "red"]  
        - Actions: ["touching", "grounded"] 
        - Image: Two blue blocks each touching a pyramid 

        ```python
        actions = [
            ['block', 'touching', 'pyramid'],
            ['block', 'touching', 'pyramid']
        ]
        ```
        *Note: The interaction occurs twice, so it appears twice in the list*  

        **Example 2**
        - Objects: ["block", "wedge"]                                                                                     
        - Properties: ["blue", "red", "upright"]                                                                          
        - Actions: ["touching", "supporting"]                                                                             
        - Image: Blue block touching red wedge, wedge supporting a block      

        ```python
        actions = [
            ['blue', 'touching', 'red'],                                                                                  
            ['wedge', 'supporting', 'block'] 
        ]
        ```

        **Analyze the image now:**
\end{lstlisting}

The following prompt is used during the experimentation to get an example suggestion from the VLM:
\begin{lstlisting}
    You are a Zendo player. Your job is to generate a new structure example to gain new knowledge about the hidden rule.
    You are given a few positive and negative examples. Each structure consists of a list of items with the format:
    "item(ID, color, shape, orientation, interaction)".
    
    The pieces can have colors: red, blue, yellow; shapes: block, wedge, pyramid; orientations: upright, upside_down, flat, cheesecake, doorstop.
    Wedges are never flat but instead can be doorstop or cheesecake, while the two other shapes can be flat but not cheesecake or doorstop.
    Interactions can be: grounded, touching(ID), pointing(ID) and on_top_of(ID), where ID is the first field of another piece, e.g. "pointing(2)" means this piece is pointing to the piece with ID 2.
    You may propose up to 7 pieces in your structure.
    The current top rule hypotheses are:
    {top_rules_str}
    
    Positive examples: First {positive_count} examples are positive, meaning they follow the hidden rule, while the next {negative_count} examples are negative, meaning they do not follow the hidden rule.
    
    Here are examples of valid answers:
    ```python
    [["item(0, red, block, upright, grounded)", "item(1, blue, wedge, doorstop, touching(0))"], 1]
    ```
    ```python
    [["item(0, yellow, pyramid, upside_down, grounded)", "item(1, blue, wedge, doorstop, on_top_of(0))", "item(2, red, block, upright, pointing(0))"], 0]
    ```
    Please ONLY return a new example and its label within a python block, in this exact format, where label is 1 for valid and 0 for invalid:
    ```python
    [["item(ID, color, shape, orientation, interaction)", ...], label]
    ```
\end{lstlisting}

\subsection{Bayesian Agent}
The Bayesian Agent uses the following prompt to get initial hypotheses:
\begin{lstlisting}
    Please list {num} possible rules about the {att} with the choices {att_choices}.
    
    Example 1:
    {example}
    Do NOT propose rules containing none such as "there is a wedge pointing at nothing/none".
    
    Task 1:
    Image is attached.
    Simple rules (Orders do NOT matter):
\end{lstlisting}
During the experimentation, we use the following prompt to get examples:
\begin{lstlisting}
    Given the rule '{h}', please give one structure that conforms with the rule and another structure that violates with the rule. 
    
    A structure has one or more pieces. Each piece should contain the following attributes: 
    - color: red, blue, yellow;
    - shape: block, wedge, pyramid;
    - orientation: upright, upside_down, flat, cheesecake, doorstop;
    - relation: grounded (whether the piece is touching the ground), touching(ID) (whether the piece is touching another piece with ID), pointing(ID) (whether the piece is pointing to another piece with ID), on_top_of(ID) (whether the piece is on top of another piece with ID).
    Wedges are never flat but instead can be doorstop or cheesecake, while the two other shapes can be flat but not cheesecake or doorstop.
    Interactions can be: grounded, touching(ID), pointing(ID) and on_top_of(ID), where ID is the first field of another piece, e.g. "pointing(2)" means this piece is pointing to the piece with ID 2.
    
    Use prolog style lists of items to represent structures. Each structure should be labeled with 1 if it conforms with the rule, and 0 if it violates the rule.
    Here is an example of a valid answer:
    [["item(0, red, block, upright, grounded)", "item(1, blue, wedge, doorstop, touching(0))"], 0]
    [["item(0, yellow, pyramid, upside_down, grounded)", "item(1, blue, wedge, doorstop, on_top_of(0))", "item(2, red, block, upright, pointing(1))"], 1]
    
    Return your answer in this exact format:
    [["item(ID, color, shape, orientation, interaction)", ...], 0]
    [["item(ID, color, shape, orientation, interaction)", ...], 1]
\end{lstlisting}
Over the course of the game, the Agent revises the initial hypotheses using the following prompt:
\begin{lstlisting}
    A structure has one or more blocks. Each block should contain the following attributes: 
    - color: red, blue, yellow;
    - shape: block, wedge, pyramid;
    - orientation: upright, upside_down, flat, cheesecake, doorstop;
    - relation: grounded (whether the piece is touching the ground), touching(ID) (whether the piece is touching another piece with ID), pointing(ID) (whether the piece is pointing to another piece with ID), on_top_of(ID) (whether the piece is on top of another piece with ID).
    Wedges can be doorstop, cheesecake or flat, while the two other shapes can be flat but not cheesecake or doorstop.
    Interactions can be: grounded, touching(ID), pointing(ID) and on_top_of(ID), where ID is the first field of another piece, e.g. "pointing(2)" means this piece is pointing to the piece with ID 2.
    
    Example of rule modifications: 
    Quantifier change: 'There must be a blue block' -> 'There are two blue blocks'
    Additional attribute: 'There must be a blue block' -> 'There must be a blue block that is upright'
    Attribute change: 'There must be a blue block' -> 'There must be a yellow block'
    These modifications are "local": only one attribute/quantifier is changed or added for each modification.
    
    Please modify the rule '{h}'. Generate {num} rules for each type of modification (Quantifier change, Additional attribute, Attribute change) so that the appended image of a structure is {text_y} a good structure (follows the rule):
    
    Make the format a numbered list (1., 2., ..., 5.) Remember that the new rules should be a "local" modification from the rule '{h}'. Do not use attribute values that are not mentioned earlier. Do not say anything other than the modified rules.
\end{lstlisting}

%% file: user_instructions.tex
\textbf{Information Sheet and Data Protection}

\textbf{Purpose and Potential Benefit.}
The purpose of this study is to examine how people learn and reason about visual rules. The results have the potential to improve our understanding of human inductive reasoning and concept learning.

\textbf{Procedure.}
After a short briefing and tutorial, participants will play a computer-based game.

\textbf{Duration.}
Participation takes approximately 20 minutes. There is no monetary compensation.

\textbf{Risks.}
Participation does not involve any risks beyond those encountered in everyday life.

\textbf{Privacy.}
No personal data is collected. We record only task-related interactions, including constructed scenes, label guesses, rule guesses, and timing information. These data are linked to an anonymous participant ID and cannot be traced back to individuals.

\textbf{Storage.}
Data are stored on university servers and internal systems in anonymised form and may be used in scientific publications.

\textbf{Voluntariness.}
Participation is voluntary. Participants may withdraw at any time without disadvantage.

\textbf{Consent.}
By continuing, participants agree to participate and consent to anonymous data storage and use.

\textbf{Contact.}
Sophia Koehler (sophia.koehler@tu-darmstadt.de).

\vspace{0.5em}
\textbf{Game Instructions: Zendo}

\textbf{Goal.}
Identify a hidden rule in \textbf{as few turns as possible}.

\textbf{Pieces.}
Scenes consist of 3D pieces varying in:
\begin{itemize}
    \item Shape: block, pyramid, wedge
    \item Color: red, blue, yellow
    \item Orientation: upright, upside-down, flat, cheesecake, doorstop
\end{itemize}
Pieces may touch, stack, or point toward each other.

\textbf{Start of the Game.}
You are shown one positive (YES) and one negative (NO) example.

\textbf{Turn Structure.}
\begin{enumerate}
    \item Build a scene to test your hypothesis.
    \item Guess whether it follows the rule (YES/NO).
    \item If correct, optionally guess the rule, if incorrect, receive a counter-example.
\end{enumerate}

\textbf{Ending.}
The game ends when the rule is found or after 30 labelled scenes.

\textbf{Tips.}
\begin{itemize}
    \item Compare YES and NO scenes.
    \item Change one feature at a time.
    \item Track previous guesses.
\end{itemize}

%% file: references.bib
@article{COOK2011341,
title = {Where science starts: Spontaneous experiments in preschoolers’ exploratory play},
journal = {Cognition},
volume = {120},
number = {3},
pages = {341-349},
year = {2011},
note = {Probabilistic models of cognitive development},
issn = {0010-0277},
author = {Claire Cook and Noah D. Goodman and Laura E. Schulz},
}

@article{oaksford1994rational,
  title={A rational analysis of the selection task as optimal data selection.},
  author={Oaksford, Mike and Chater, Nick},
  journal={Psychological review},
  volume={101},
  number={4},
  pages={608},
  year={1994},
  publisher={American Psychological Association}
}

@article{lange2015just,
  title={" Just another tool for online studies”(JATOS): An easy solution for setup and management of web servers supporting online studies},
  author={Lange, Kristian and K{\"u}hn, Simone and Filevich, Elisa},
  journal={PloS one},
  volume={10},
  number={6},
  pages={e0130834},
  year={2015},
  publisher={Public Library of Science San Francisco, CA USA}
}

@manual{xcolor,
	title      = {The \texttt{xcolor} package},
	subtitle   = {Driver-independent color extensions for \LaTeX{} and pdfLaTeX},
	author     = {Uwe Kern and {\LaTeX{} Project}},
	date       = {2024-09-29},
	version    = {3.02},
	url        = {http://mirrors.ctan.org/macros/latex/contrib/xcolor/xcolor.pdf}
}

@article{shindo2023alpha,
  title={$\alpha$ ilp: thinking visual scenes as differentiable logic programs},
  author={Shindo, Hikaru and Pfanschilling, Viktor and Dhami, Devendra Singh and Kersting, Kristian},
  journal={Machine Learning},
  volume={112},
  number={5},
  pages={1465--1497},
  year={2023},
  publisher={Springer}
}

@inproceedings{marsili2025visual,
  title={Visual Agentic AI for Spatial Reasoning with a Dynamic API},
  author={Marsili, Damiano and Agrawal, Rohun and Yue, Yisong and Gkioxari, Georgia},
  booktitle={2025 IEEE/CVF Conference on Computer Vision and Pattern Recognition (CVPR)},
  pages={19446--19455},
  year={2025},
  organization={IEEE}
}

@inproceedings{wust2024pix2code,
  title={Pix2Code: Learning to Compose Neural Visual Concepts as Programs},
  author={W{\"u}st, Antonia and Stammer, Wolfgang and Delfosse, Quentin and Dhami, Devendra Singh and Kersting, Kristian},
  booktitle={Uncertainty in Artificial Intelligence},
  pages={3829--3852},
  year={2024},
  organization={PMLR}
}

@article{ellis2023dreamcoder,
  title={DreamCoder: growing generalizable, interpretable knowledge with wake--sleep Bayesian program learning},
  author={Ellis, Kevin and Wong, Lionel and Nye, Maxwell and Sable-Meyer, Mathias and Cary, Luc and Anaya Pozo, Lore and Hewitt, Luke and Solar-Lezama, Armando and Tenenbaum, Joshua B},
  journal={Philosophical Transactions of the Royal Society A},
  volume={381},
  number={2251},
  pages={20220050},
  year={2023},
  publisher={The Royal Society}
}

@article{piriyakulkij2024doing,
  title={Doing experiments and revising rules with natural language and probabilistic reasoning},
  author={Piriyakulkij, Top and Langenfeld, Cassidy and Le, Tuan Anh and Ellis, Kevin},
  journal={Advances in Neural Information Processing Systems},
  volume={37},
  pages={53102--53137},
  year={2024}
}

@inproceedings{bramley2018grounding,
  title={Grounding compositional hypothesis generation in specific instances},
  author={Bramley, Neil R and Rothe, Anslem and Tenenbaum, Joshua B and Xu, Fei and Gureckis, Todd M},
  booktitle={Proceedings of the Annual Meeting of the Cognitive Science Society},
  volume={40},
  year={2018}
}

@article{chollet2019measure,
  title={On the measure of intelligence},
  author={Chollet, Fran{\c{c}}ois},
  journal={arXiv preprint arXiv:1911.01547},
  year={2019}
}

@inproceedings{fijalkow2022scaling,
  title={Scaling neural program synthesis with distribution-based search},
  author={Fijalkow, Nathana{\"e}l and Lagarde, Guillaume and Matricon, Th{\'e}o and Ellis, Kevin and Ohlmann, Pierre and Potta, Akarsh Nayan},
  booktitle={Proceedings of the AAAI Conference on Artificial Intelligence},
  volume={36},
  number={6},
  pages={6623--6630},
  year={2022}
}

@article{johnson2023image,
  title={Image Analysis through the lens of ChatGPT-4},
  author={Johnson, Olanrewaju Victor and Alyasiri, Osamah Mohammed and Akhtom, Dua’a and Johnson, Olabisi Esher},
  journal={Journal of Applied Artificial Intelligence},
  volume={4},
  number={2},
  pages={31--46},
  year={2023}
}

@article{wust2025synthesizing,
  title={Synthesizing Visual Concepts as Vision-Language Programs},
  author={W{\"u}st, Antonia and Stammer, Wolfgang and Shindo, Hikaru and Dhami, Devendra Singh and Helff, Lukas and Kersting, Kristian},
  journal = {IEEE/CVF Conference on Computer Vision and Pattern Recognition},
  year      = {2026}
}

@article{beger2025ai,
  title={Do AI Models Perform Human-like Abstract Reasoning Across Modalities?},
  author={Beger, Claas and Yi, Ryan and Fu, Shuhao and Denton, Kaleda and Moskvichev, Arseny and Tsai, Sarah W and Rajamanickam, Sivasankaran and Mitchell, Melanie},
  journal={arXiv preprint arXiv:2510.02125},
  year={2025}
}

@inproceedings{wang2023hypothesis,
  title={Hypothesis Search: Inductive Reasoning with Language Models},
  author={Wang, Ruocheng and Zelikman, Eric and Poesia, Gabriel and Pu, Yewen and Haber, Nick and Goodman, Noah},
  booktitle={The Twelfth International Conference on Learning Representations},
year={2023}
}

@book{hofstadter1995fluid,
  title={Fluid concepts and creative analogies: Computer models of the fundamental mechanisms of thought.},
  author={Hofstadter, Douglas R},
  year={1995},
  publisher={Basic books}
}

@article{Feist1998ThePO,
  title={The Psychology of Science: Review and Integration of a Nascent Discipline},
  author={Gregory J. Feist and Michael E. Gorman},
  journal={Review of General Psychology},
  year={1998},
  volume={2},
  pages={3 - 47}
}

@inproceedings{Zhou2025PhysVLMAVRAV,
  title={PhysVLM-AVR: Active Visual Reasoning for Multimodal Large Language Models in Physical Environments},
  author={Zhou, Weijie and Xiong, Xuantang and Peng, Yi and Tao, Manli and Zhao, Chaoyang and Dong, Honghui and Tang, Ming and Wang, Jinqiao},
  booktitle={The Thirty-ninth Annual Conference on Neural Information Processing Systems},
  year = {2025}
}

@inproceedings{helff2026slr,
  title     = {{SLR}: Automated Synthesis for Scalable Logical Reasoning},
  author    = {Helff, Lukas and Omar, Ahmad and Friedrich, Felix and W{\"u}st, Antonia and Shindo, Hikaru and Mitchell, Rupert and Woydt, Tim and Schramowski, Patrick and Stammer, Wolfgang and Kersting, Kristian},
  booktitle = {Proceedings of the 64th Annual Meeting of the Association for Computational Linguistics (ACL 2026)},
  year      = {2026},
  publisher = {Association for Computational Linguistics},
}

@article{cerrato2026science,
  title={Science-Gym: a simple testbed for AI-driven scientific discovery},
  author={Cerrato, Mattia and Baur, Lennart and Brugger, Jannis and Shumaly, Sajjad and Schmitt, Nicholas and Finkelstein, Edward and Jukic, Selina and M{\"u}nzel, Lars and Paul, Felix Peter and Pfannes, Pascal and others},
  journal={Machine Learning},
  volume={115},
  number={1},
  pages={16},
  year={2026},
  publisher={Springer}
}

@inproceedings{
li2025combining,
title={Combining Induction and Transduction for Abstract Reasoning},
author={Wen-Ding Li and Keya Hu and Carter Larsen and Yuqing Wu and Simon Alford and Caleb Woo and Spencer M. Dunn and Hao Tang and Wei-Long Zheng and Yewen Pu and Kevin Ellis},
booktitle={The Thirteenth International Conference on Learning Representations},
year={2025}
}

@article{xie2025far,
  title={How far are AI scientists from changing the world?},
  author={Xie, Qiujie and Weng, Yixuan and Zhu, Minjun and Shen, Fuchen and Huang, Shulin and Lin, Zhen and Zhou, Jiahui and Mao, Zilan and Yang, Zijie and Yang, Linyi and others},
  journal={arXiv preprint arXiv:2507.23276},
  year={2025}
}

@article{king2009robot,
  title={The Robot Scientist Adam},
  author={King, Ross D and Rowland, Jem and Aubrey, Wayne and Liakata, Maria and Markham, Magdalena and Soldatova, Larisa N and Whelan, Ken E and Clare, Amanda and Young, Mike and Sparkes, Andrew and others},
  journal={Computer},
  volume={42},
  number={8},
  pages={46--54},
  year={2009},
  publisher={IEEE Computer Society Press Los Alamitos, CA, USA}
}

@article{Lu2024TheASA,
  title={The AI Scientist: Towards Fully Automated Open-Ended Scientific Discovery},
  author={Chris Lu and Cong Lu and R. Lange and Jakob Foerster and Jeff Clune and David Ha},
  journal={ArXiv},
  year={2024},
  volume={abs/2408.06292},
}

@inproceedings{wubongard,
  author    = {Wu, Rujie and Ma, Xiaojian and Zhang, Zhenliang and Wang, Wei and Li, Qing and Zhu, Song-Chun and Wang, Yizhou},
  title     = {{Bongard-OpenWorld}: Few-Shot Reasoning for Free-form Visual Concepts in the Real World},
  booktitle = {International Conference on Learning Representations},
  year      = {2024}
}

@inproceedings{jiang2022bongard,
  author    = {Jiang, Huaizu and Ma, Xiaojian and Nie, Weili and Yu, Zhiding and Zhu, Yuke and Anandkumar, Anima},
  title     = {{Bongard-HOI}: Benchmarking Few-Shot Visual Reasoning for Human-Object Interactions},
  booktitle = {IEEE/CVF Conference on Computer Vision and Pattern Recognition},
  year      = {2022}
}

@inproceedings{wuest2025bongard,
  author    = {W{\"u}st, Antonia and Tobiasch, Tim and Helff, Lukas and Ibs, Inga and Stammer, Wolfgang and Dhami, Devendra Singh and Rothkopf, Constantin A. and Kersting, Kristian},
  title     = {Bongard in Wonderland: Visual Puzzles that Still Make {AI} Go Mad?},
  booktitle = {International Conference on Machine Learning},
  year      = {2025}
}

@book{bongard_pattern_1970,
	title = {Pattern {Recognition}},
	publisher = {Spartan Books},
	author = {Bongard, Mikhail M.},
    editor = {Hawkins, J.K.},
	year = {1970},
}

@inproceedings{zhang2019raven,
  author       = {Chi Zhang and
                  Feng Gao and
                  Baoxiong Jia and
                  Yixin Zhu and
                  Song{-}Chun Zhu},
  title        = {{RAVEN:} {A} Dataset for Relational and Analogical Visual REasoNing},
  booktitle    = {Conference on Computer Vision and Pattern Recognition (CVPR)},
  year         = {2019},
}

@article{steinmann2025object,
                title={Object Centric Concept Bottlenecks},
                author={Steinmann, David and Stammer, Wolfgang and W{\"u}st, Antonia and Kersting, Kristian},
                journal={Advances in Neural Information Processing Systems (NeurIPS)},
                year={2025}
                }

@article{warrier2025benchmarking,
  title={Benchmarking World-Model Learning},
  author={Warrier, Archana and Nguyen, Dat and Naim, Michelangelo and Jain, Moksh and Liang, Yichao and Schroeder, Karen and Yang, Cambridge and Tenenbaum, Joshua B and Vollmer, Sebastian and Ellis, Kevin and others},
  journal={arXiv preprint arXiv:2510.19788},
  year={2025}
}

@article{arcagi3_2026,
  title        = {{ARC-AGI-3}: A New Challenge for Frontier Agentic Intelligence},
  author       = {{ARC Prize Foundation}},
  journal      = {arXiv preprint arXiv:2603.24621},
  year         = {2026}
}

@article{waytowich2024atari,
  title={Atari-GPT: Benchmarking Multimodal Large Language Models as Low-Level Policies in Atari Games},
  author={Waytowich, Nicholas R and White, Devin and Sunbeam, MD and Goecks, Vinicius G},
  journal={arXiv preprint arXiv:2408.15950},
  year={2024}
}

@article{kuttler2020nethack,
  title={The nethack learning environment},
  author={K{\"u}ttler, Heinrich and Nardelli, Nantas and Miller, Alexander and Raileanu, Roberta and Selvatici, Marco and Grefenstette, Edward and Rockt{\"a}schel, Tim},
  journal={Advances in Neural Information Processing Systems},
  volume={33},
  pages={7671--7684},
  year={2020}
}

@inproceedings{malkinski2025reasoning,
  title={Reasoning Limitations of Multimodal Large Language Models. A case study of Bongard Problems},
  author={Ma{\l}ki{\'n}ski, Miko{\l}aj and Pawlonka, Szymon and Ma{\'n}dziuk, Jacek},
  booktitle={Forty-second International Conference on Machine Learning},
year={2025}
}

@article{Pawlonka25bongardrwr,
  title={Bongard-rwr+: Real-world Representations of Fine-grained Concepts in Bongard Problems},
  author={Pawlonka, Szymon and Ma{\l}ki{\'n}ski, Miko{\l}aj and Ma{\'n}dziuk, Jacek},
  journal={arXiv preprint arXiv:2508.12026},
  year={2025}
}

@article{nie2020bongard,
  title={Bongard-logo: A new benchmark for human-level concept learning and reasoning},
  author={Nie, Weili and Yu, Zhiding and Mao, Lei and Patel, Ankit B and Zhu, Yuke and Anandkumar, Anima},
  journal={Advances in neural information processing systems},
  volume={33},
  pages={16468--16480},
  year={2020}
}

@article{fan2022minedojo,
  title={Minedojo: Building open-ended embodied agents with internet-scale knowledge},
  author={Fan, Linxi and Wang, Guanzhi and Jiang, Yunfan and Mandlekar, Ajay and Yang, Yuncong and Zhu, Haoyi and Tang, Andrew and Huang, De-An and Zhu, Yuke and Anandkumar, Anima},
  journal={Advances in Neural Information Processing Systems},
  volume={35},
  pages={18343--18362},
  year={2022}
}

@book{pearl2009causality,
  title={Causality},
  author={Pearl, Judea},
  year={2009},
  publisher={Cambridge university press}
}

@book{pearl2018book,
  title={The book of why: The new science of cause and effect},
  author={Pearl, Judea},
  year={2018},
  publisher={Basic Books}
}

@article{Gopnik2000DetectingBH,
  title={Detecting blickets: how young children use information about novel causal powers in categorization and induction.},
  author={Alison Gopnik and David M. Sobel},
  journal={Child development},
  year={2000},
  volume={71 5},
  pages={
          1205-22
        },
  url={https://api.semanticscholar.org/CorpusID:17245630}
}

@inproceedings{zhang2021acre,
  title={Acre: Abstract causal reasoning beyond covariation},
  author={Zhang, Chi and Jia, Baoxiong and Edmonds, Mark and Zhu, Song-Chun and Zhu, Yixin},
  booktitle={2021 IEEE/CVF Conference on Computer Vision and Pattern Recognition (CVPR)},
  pages={10638--10648},
  year={2021},
  organization={IEEE}
}

@article{hocquette2024learning,
  title={Learning big logical rules by joining small rules},
  author={Hocquette, C{\'e}line and Niskanen, Andreas and Morel, Rolf and J{\"a}rvisalo, Matti and Cropper, Andrew},
  journal={arXiv preprint arXiv:2401.16215},
  year={2024}
}

@inproceedings{cropper2023learning,
  title={Learning logic programs by combining programs},
  author={Cropper, Andrew and Hocquette, C{\'e}line},
  booktitle={ECAI 2023: 26th European Conference on Artificial Intelligence, September 30--October 4, 2023, Krak{\'o}w, Poland--Including 12th Conference on Prestigious Applications of Intelligent Systems (PAIS 2023)},
  pages={501--508},
  year={2023},
  organization={SAGE Publications 1 Oliver's Yard, 55 City Road, London, EC1Y 1SP}
}
